\documentclass[]{fairmeta}
\usepackage{xspace}

\makeatletter
\DeclareRobustCommand\onedot{\futurelet\@let@token\@onedot}
\def\@onedot{\ifx\@let@token.\else.\null\fi\xspace}

\def\eg{\emph{e.g}\onedot} 
\def\ie{\emph{i.e}\onedot}

\makeatother

\usepackage{wrapfig}
\usepackage{colortbl}
\usepackage{color, xcolor} 
\usepackage{listings}
\usepackage{enumitem}

\def\ourmodel{\texttt{OneStory}}

\definecolor{cvprblue}{rgb}{0.21,0.49,0.74}
\definecolor{row}{RGB}{235, 245, 251}
\usepackage{makecell}
\usepackage{wrapfig}
\usepackage{amssymb}
\usepackage{graphicx}   
\usepackage{subcaption} 
\usepackage{booktabs}   
\usepackage{multirow}   
\newcolumntype{C}[1]{>{\centering\arraybackslash}m{#1}}
\newcommand{\projectpage}{\href{https://zhaochongan.github.io/projects/OneStory}{\textcolor{magenta}{Project Page}}\xspace}

\title{\textcolor{metablue}{OneStory}: Coherent Multi-Shot Video Generation with Adaptive Memory}

\author[1,2]{Zhaochong An}
\author[1]{Menglin Jia}
\author[1]{Haonan Qiu}
\author[1]{Zijian Zhou}
\author[1]{Xiaoke Huang}
\author[1]{Zhiheng Liu}
\author[1]{Weiming Ren}
\author[1]{Kumara Kahatapitiya}
\author[1]{Ding Liu}
\author[1]{Sen He}
\author[1]{Chenyang Zhang}
\author[1]{Tao Xiang}
\author[1]{Fanny Yang}
\author[2]{Serge Belongie}
\author[1,*]{Tian Xie}

\affiliation[1]{Meta AI}
\affiliation[2]{University of Copenhagen}

\contribution[*]{Project lead}

\abstract{Storytelling in real-world videos often unfolds through multiple shots—discontinuous yet semantically connected clips that together convey a coherent narrative.
However, existing multi-shot video generation (MSV) methods struggle to effectively model long-range cross-shot context, as they rely on limited temporal windows or single keyframe conditioning, leading to degraded performance under complex narratives.
In this work, we propose \textbf{\ourmodel}, enabling global yet compact cross-shot context modeling for consistent and scalable narrative generation.
\ourmodel~reformulates MSV as a next-shot generation task, enabling autoregressive shot synthesis while leveraging pretrained image-to-video (I2V) models for strong visual conditioning.
We introduce two key modules: a \textbf{Frame Selection} module that constructs a semantically-relevant global memory based on informative frames from prior shots, and an \textbf{Adaptive Conditioner}
that performs importance-guided patchification to generate compact context for direct conditioning.
We further curate a high-quality multi-shot dataset with referential captions to mirror real-world storytelling patterns, and design effective training strategies under the next-shot paradigm.
Finetuned from a pretrained I2V model on our curated 60K dataset, \ourmodel\ achieves state-of-the-art narrative coherence across diverse and complex scenes in both text- and image-conditioned settings, enabling controllable and immersive long-form video storytelling.}

\correspondence{\email{zhan@di.ku.dk}}
\metadata[Project page]{\url{https://zhaochongan.github.io/projects/OneStory}}

\begin{document}

\maketitle

\section{Introduction}
\label{sec:intro}

\begin{figure*}[t!]
\centering
\includegraphics[width=\linewidth]{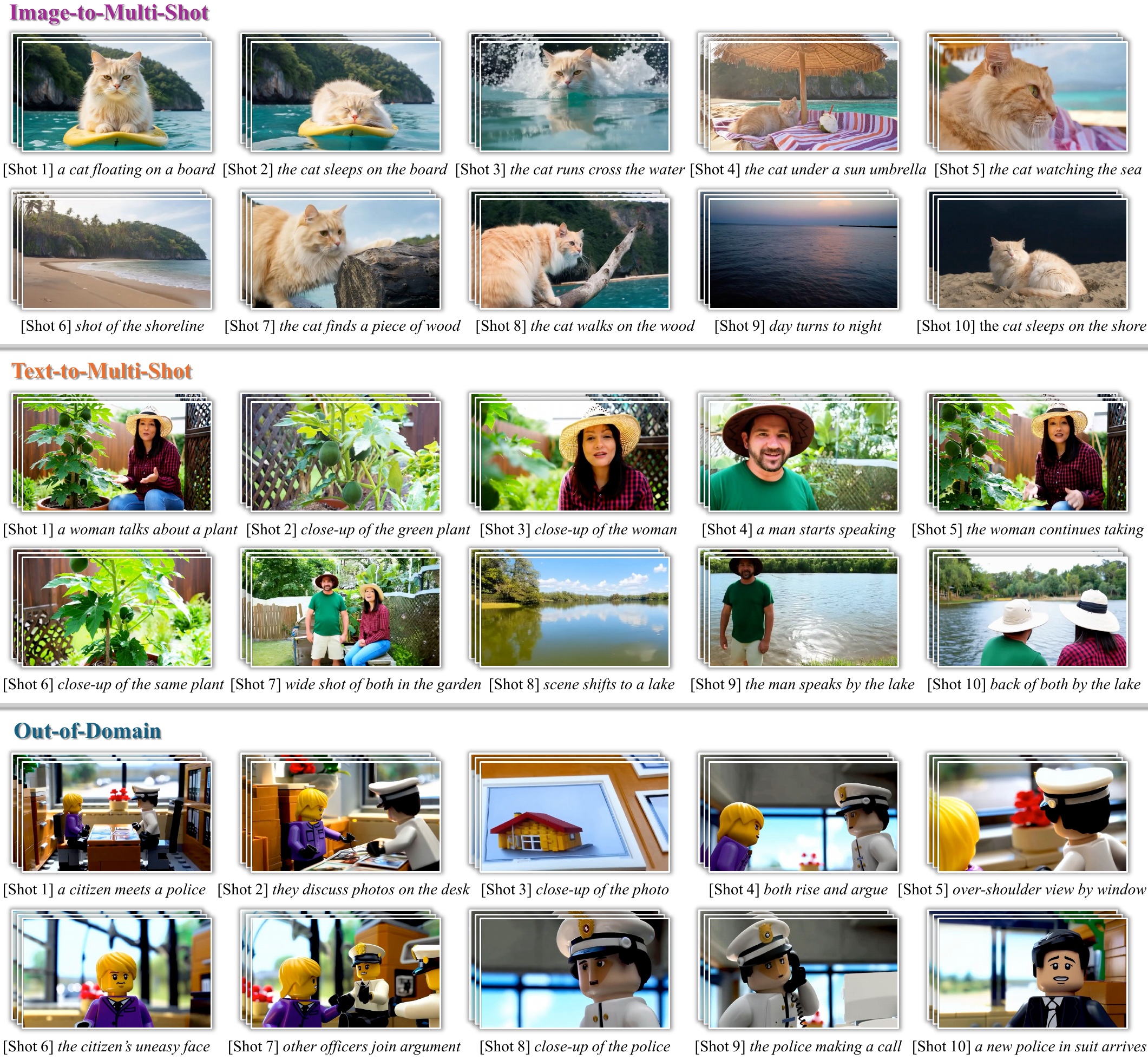}
    \caption{\textbf{Coherent multi-shot generations with \ourmodel.}
Each example shows 10-shots of a minute-long video. \ourmodel~handles both image-to-multi-shot (\textit{top}) and text-to-multi-shot (\textit{middle}) generation within the same model, and generalizes well to out-of-domain scenes (\textit{bottom}). It maintains consistent characters and environments while faithfully following complex and evolving prompts to produce coherent long-form narratives. A representative segment of each prompt is given with the corresponding shot. \textit{We recommend referring to our \projectpage for better visualization.}
}
    \label{fig:teaser}
\end{figure*}

Recent advances in diffusion transformers~\citep{peebles2023scalable} have greatly advanced video generation~\citep{yu2025context,chuwan,song2025history,wang2025cinemaster}, achieving impressive visual fidelity.
Despite this success, current models remain largely confined to producing a \emph{single continuous scene}, ignoring long-range narrative modeling. 
In contrast, real-world applications~\citep{xing2025motioncanvas,gao2025seedance,qiu2025animeshooter} demand \emph{multi-shot videos}: a sequence of shots that together convey a coherent storyline.
Consequently, \emph{multi-shot video generation (MSV)} is emerging as a critical research direction~\citep{wang2024dreamrunner,lu2025skald,jiang2025lovic,atzmon2024multi}.

Compared to single-shot generation, MSV is inherently more challenging as it requires both \emph{narrative consistency} and \emph{spatio-temporal reasoning} across discontinuous scenes~\citep{liu2025shotbench,wang2025cinetechbench}.
First, consistent narrative entities, such as characters and environments, must persist even when intermittently off-screen~\citep{guo2025long,song2025lavieid}.
Second, as consecutive shots may vary in terms of time, location, and viewpoint, the model must discern which aspects should remain invariant (\eg, identity, scene layout) and which should evolve (\eg, camera angles, actions)~\citep{lin2025towards,liao2025thinking,chen2025talkcuts}.
In essence, the core difficulty of MSV lies in effectively exploiting and maintaining the \emph{long-term cross-shot context}.

Based on how the cross-shot context is modeled, existing approaches can be categorized into two paradigms.  
\textbf{(1) Fixed-window attention} paradigm~\citep{wu2025cinetrans,kara2025shotadapter,qi2025mask,guo2025long} computes attention over several shots within a fixed temporal window, by applying caption-to-shot attention masks~\citep{kara2025shotadapter,qi2025mask} or direct long context tuning~\citep{guo2025long}.
However, due to the fixed window size, older shots are discarded as the window slides forward, leading to inevitable memory loss and inconsistency beyond the window.  
\textbf{(2) Keyframe conditioning} paradigm~\citep{he2025cut2next,zhao2024moviedreamer,long2024videostudio,xiao2025captain} generates a keyframe for each shot and expands it into a full clip using image-to-video (I2V) models before concatenation.
However, such multi-stage pipelines restrict the cross-shot context to a single image, limiting the propagation of complex narrative cues and resulting in weak storyline adherence.

To this end, we propose \textbf{\ourmodel}, an effective framework that overcomes the context modeling limitations in prior work.
First, we reformulate MSV as a \emph{next-shot generation} task, to enable autoregressive shot synthesis and leverage the strong visual conditioning capability of pretrained I2V models.
Second, inspired by varying correlations across shots,
we introduce a \textbf{Frame Selection} module that identifies a sparse set of semantically-relevant frames across all prior shots, effectively mitigating memory loss and recovering long-range context.
Third,  
we design an \textbf{Adaptive Conditioner} 
that patchifies the selected context dynamically and injects directly into the generator, providing efficient and expressive conditioning.
Unlike prior works~\citep{gu2025long,zhang2025packing} that rely on a fixed temporal ordering, our conditioner compresses a set of disconnected frames adaptively based on their importance.
Together, our model enables an \emph{global yet compact} cross-shot context, supporting consistent and scalable story generation.

Beyond model design, 
we carefully curate a high-quality dataset of $\sim$60K multi-shot videos through a three-step pipeline comprising \emph{shot detection}, \emph{two-stage captioning}, and \emph{quality filtering} (see~\Cref{sec:data_process}).
This dataset reflects realistic storytelling where captions are provided shot by shot in a referential narrative flow, providing a better flexibility to evolve, without the need of having a separate global script~\citep{guo2025long,wu2025cinetrans}.
We also propose effective training strategies under the new MSV formulation,
including unified three-shot training and a progressive coupling scheme, to facilitate end-to-end optimization and narrative coherence.

Built upon a pretrained I2V model and finetuned on our curated dataset, \ourmodel~achieves superior performance across diverse and complex narratives.
As shown in~\Cref{fig:teaser}, it generates minute-scale, ten-shot videos with strong visual consistency and narrative adherence. It supports both image-to-multi-shot (\textit{top}) and text-to-multi-shot (\textit{middle}), while also generalizing to out-of-domain scenes (\textit{bottom}) despite trained only on human-centric data.
The coherence, flexibility, and scalability of \ourmodel~make it well-suited for real-world creative applications, paving the way for immersive multi-shot storytelling.

\vspace{9pt}
\section{Related Works}
\label{sec:related_work}
\vspace{3pt}
\noindent\textbf{Single-shot Video Generation.}
Single-shot video generation primarily includes \textit{text-to-video (T2V)} and \textit{image-to-video (I2V)} models. 
Early T2V models~\citep{zhang2024show,chen2024videocrafter2,wang2024lavie,luo2023videofusion,zeng2024make,bar2024lumiere} extended text-to-image diffusion architectures~\citep{rombach2022high} with temporal modules, while the emergence of Diffusion Transformers (DiT)~\citep{peebles2023scalable} enabled unified spatial–temporal modeling~\citep{brooks2024video,polyak2024movie,kling,hacohen2024ltx} through transformer-based designs.
Large-scale T2V models such as Wan~\citep{wan2025wan}, HunyuanVideo~\citep{kong2024hunyuanvideo}, CogVideoX~\citep{yang2024cogvideox}, and Mochi~\citep{genmo2024mochi} further boost fidelity by leveraging billion-scale datasets.
In parallel, I2V models~\citep{chen2023seine,chen2024livephoto,shi2025motionstone,hu2024animate,lin2025stiv,huang2025conceptmaster,jiang2024videobooth} animate static images with textual conditioning, offering stronger visual realism and controllability~\citep{guo2025keyframe,wang2025frame,li2025realcam}.  
Despite these advances, these methods remain confined to \emph{single-shot} generation and thus fall short for real-world storytelling, motivating the exploration toward multi-shot video generation.

\vspace{3pt}
\noindent\textbf{Multi-shot Video Generation.}
Recent efforts in MSV primarily follow two paradigms for modeling cross-shot context:  
\textit{(i) Fixed-window attention.}  
These methods~\citep{wu2025cinetrans,bansal2024talc,qi2025mask,kara2025shotadapter,wei2025mocha,guo2025long,cai2025mixture,jia2025moga,meng2025holocine,wang2025echoshot,wu2025mind} compute attention across multiple shots within a bounded temporal window.  
Mask$^2$DiT~\citep{qi2025mask} modifies attention masks to enforce caption–shot alignment, while LCT~\citep{guo2025long} augments MMDiT~\citep{esser2024scaling} to encode multi-shot structure.  
However, their finite window inevitably discards earlier shots as the window slides forward, leading to memory loss and narrative inconsistency.
\textit{(ii) Keyframe conditioning.}  
Another line of work~\citep{xie2024dreamfactory,zhao2024moviedreamer,zheng2024videogen,hu2024storyagent,xiao2025captain,he2025cut2next,zhang2025shouldershot} decomposes multi-shot generation into subproblems by synthesizing a keyframe (or reference image~\citep{long2024videostudio}) for each shot and expanding it into a full clip with an I2V model~\citep{xing2024dynamicrafter,seawead2025seaweed}.
Captain Cinema~\citep{xiao2025captain}, for example, fine-tunes a text-to-image model~\citep{flux} for identity persistence.  
However, relying on only one keyframe per shot limits cross-shot context, hindering the propagation of complex narrative information and weakening storyline consistency.
These limitations motivate our proposed \ourmodel, which models a \emph{global yet compact cross-shot context} for consistent and scalable narrative generation.

\section{High-quality Multi-shot Video Dataset}
\label{sec:data_process}

We define a high-quality multi-shot video as one maintaining a consistent theme across shots with coherent narrative progression, rather than a concatenation of unrelated clips.
Regarding captions, existing MSV methods~\citep{guo2025long,wu2025cinetrans,cai2025mixture} typically rely on structured global prompts describing the overall storyline, characters, and environment, supplemented by per-shot prompts for local details.  
While such global scripts provide the model with overarching guidance, they restrict how subsequent shots can evolve beyond the predefined storyline.  
In contrast, we do not rely on a global script and rather construct shot-level captions with referential narrative flow derived from preceding shots, offering greater narrative flexibility and reflecting real-world storytelling, where shots evolve naturally from prior context.

\begin{wrapfigure}{r}{0.5\textwidth}
    \vspace{-10pt}
    \centering
    \includegraphics[width=\linewidth]{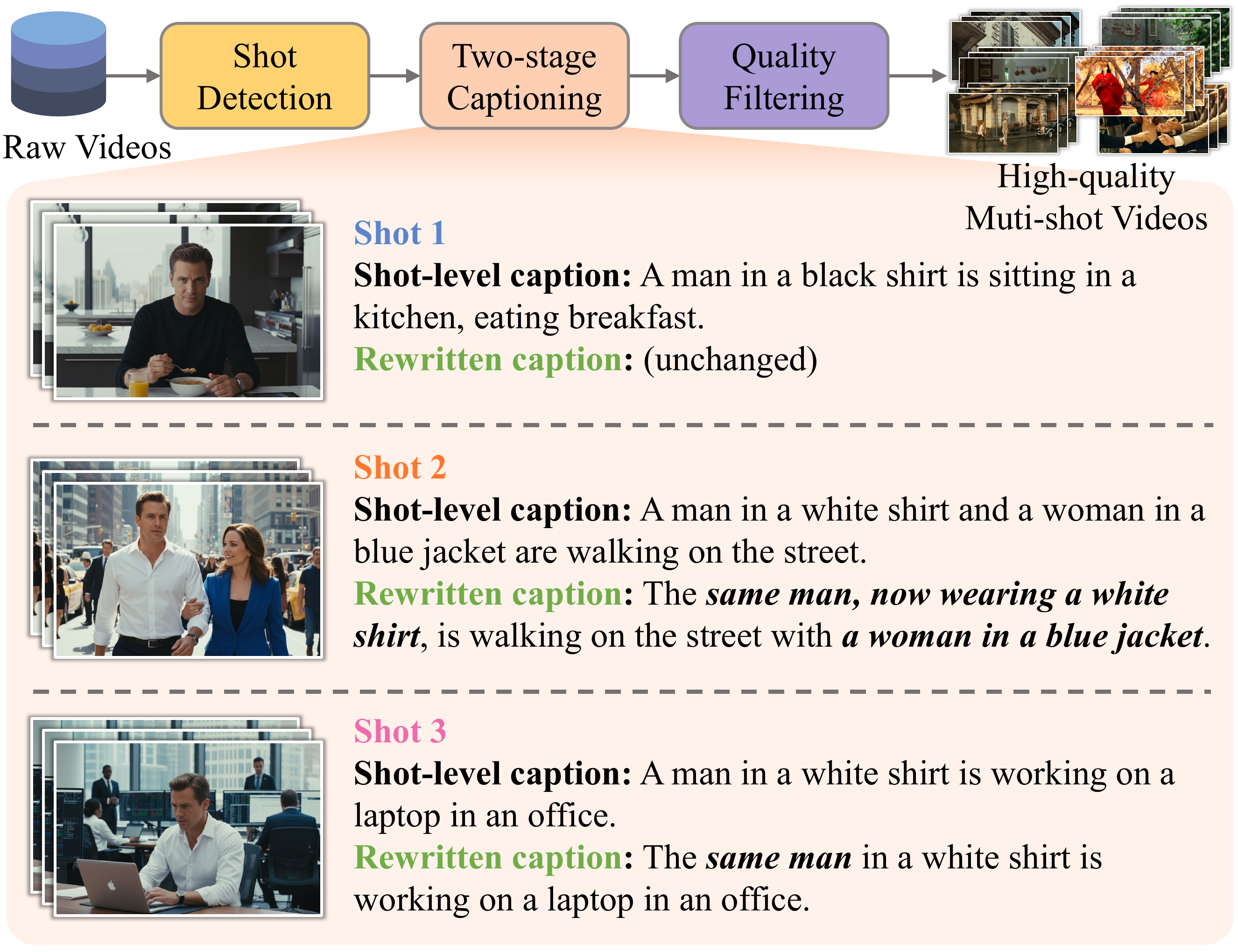}
    \caption{
\textbf{Multi-shot video data curation pipeline.}
From raw videos, we obtain high-quality multi-shot sequences via three steps:
(\textit{i}) \emph{Shot detection},
(\textit{ii}) \emph{Two-stage captioning}, and
(\textit{iii}) \emph{Quality filtering}.
In the second stage, each shot is first captioned independently and then \emph{rewritten into referential form} based on preceding shots.
Unlike prior datasets, no global captions are used, and only shot-level captions with progressive narrative flow are retained to ensure flexibility, while reflecting real-world storytelling.
    }
\label{fig:data_example}
\end{wrapfigure}

As illustrated in~\Cref{fig:data_example}, our dataset is built from videos under research copyright, focusing on human-centric activities.  
We first apply TransNetV2~\citep{soucek2024transnet} to detect shot boundaries and retain videos containing at least two shots.  
Next, we use a vision-language model~\citep{llama4,bai2025qwen2,yuan2025tarsier2} for shot-level captioning in two stages:  
(i) captioning each shot independently, then
(ii) rewriting subsequent captions based on the frames and caption of the previous shot, to introduce referential expressions (\eg, “the same man”) and describe scene/object variations.
This approach ensures contextual linkage and smooth narrative flow across shots.
After captioning, we perform multi-stage filtering to ensure quality.  
We apply keyword filters to remove videos with undesirable content.  
Then, we use feature-based filters, \ie, CLIP~\citep{radford2021learning} and SigLIP2~\citep{tschannen2025siglip}, to eliminate videos with completely irrelevant transitions, and DINOv2~\citep{oquab2023dinov2} to discard overly similar shots.  
The resulting dataset contains approximately \textbf{60K high-quality multi-shot videos} (50K two-shot and 10K three-shot), each exhibiting narrative continuity, forming a solid foundation for training multi-shot generation models.

\section{Method}
\label{sec:method}

\subsection{Task Formulation}
\label{sec:task}

Let an $N$-shot video be denoted as $V=\{S_1,\dots,S_N\}$, where each shot $S_i$ contains $K$ frames $S_i=\{f_i^1,\dots,f_i^K\}$ with spatial resolution $H\times W$.  
Each shot is paired with a caption $C_i$ that explicitly references prior shots, as detailed in~\Cref{sec:data_process}.  
Given captions $\{C_i\}_{i=1}^N$ and an optional starting image $I$ as the first-frame condition, a multi-shot video generation (MSV) model $\mathcal{G}$ aims to synthesize $V$ that faithfully follows the narrative while preserving visual consistency.

\begin{figure*}
    \centering
    \includegraphics[width=\linewidth]{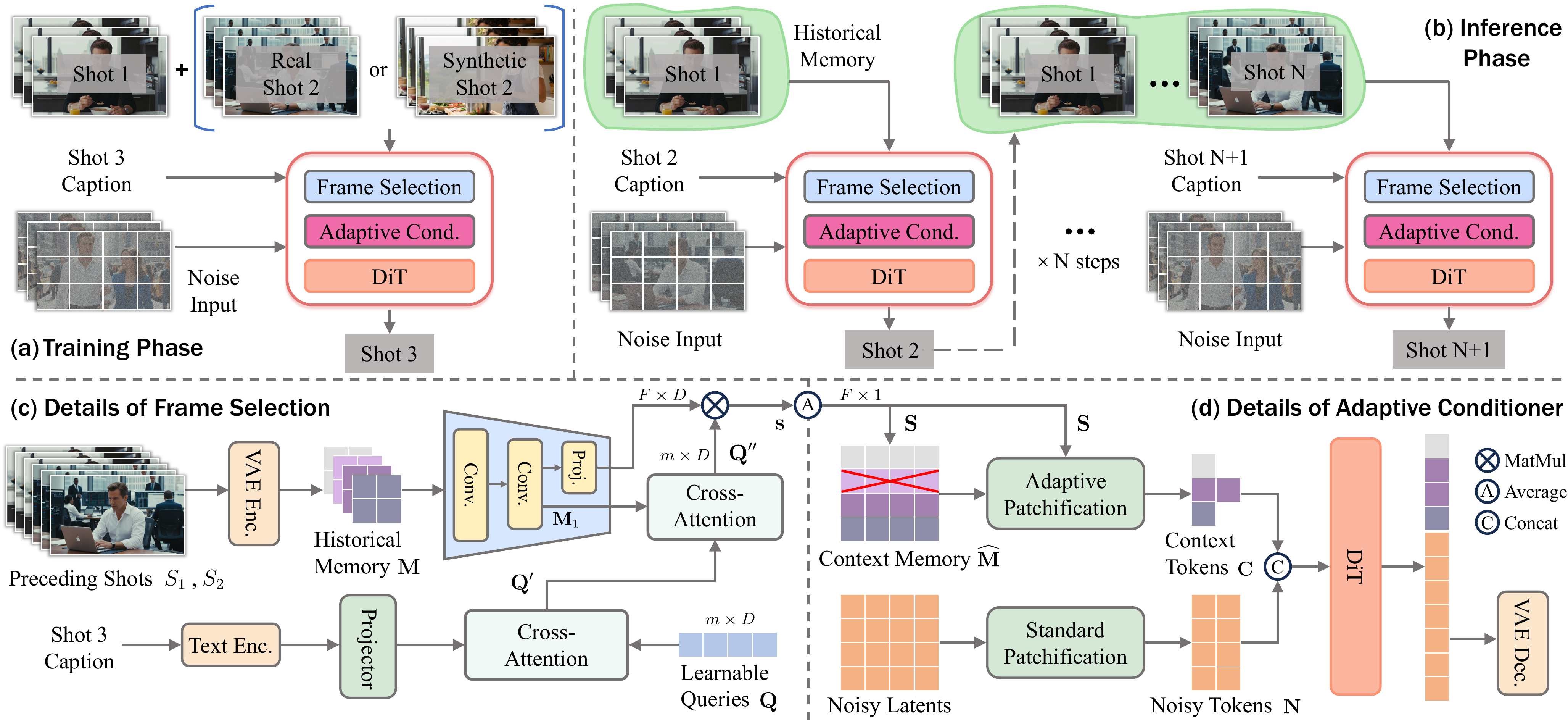}
    \caption{
\textbf{Overview of the proposed~\ourmodel.}
Our model reframes multi-shot video generation (MSV) as a \textit{next-shot generation} task.
(a)~During training, the model learns to generate the final shot conditioned on the preceding two; when only two shots are available, we inflate with a synthetic shot to enable unified three-shot training.
(b)~At inference, it maintains a memory bank of past shots and generates multi-shot videos autoregressively.
The model is comprised of two key components: (c)~a \textit{Frame Selection} module that selects semantically-relevant frames from preceding shots to construct a global context, and (d)~an \textit{Adaptive Conditioner} that dynamically compresses the selected context and injects it directly into the generator for efficient conditioning.
Together, \ourmodel~realizes adaptive memory modeling, enabling global yet compact cross-shot context for coherent narrative generation.
}
    \label{fig:method}
\end{figure*}

To enable autoregressive shot generation and leverage the strong visual conditioning capabilities of pretrained image-to-video (I2V) models, we reformulate MSV as a \emph{next-shot generation} task:
\begin{equation}
\label{eq:gen}
S_i = \mathcal{G}\big(\mathcal{E},\, \{S_j\}_{j=1}^{i-1},\, \mathcal{T},\, C_i\big),
\end{equation}
where $\mathcal{E}$ is a 3D VAE encoder~\citep{polyak2024movie,wan2025wan} that maps each shot $S_i$ into latent features $\mathbf{z}_i \in \mathbb{R}^{\frac{K}{f_t}\times \frac{H}{f_s}\times \frac{W}{f_s}\times D_v}$, with $f_t$ and $f_s$ denoting temporal and spatial compression factors, and $D_v$ the latent dimension.  
The text encoder $\mathcal{T}$ encodes caption $C_i$ into $T$ tokens $\mathbf{t}_i = \mathcal{T}(C_i) \in \mathbb{R}^{T\times D_t}$ of dimension $D_t$.
Under this formulation, our model \textbf{\ourmodel}, initialized from a pretrained I2V model, achieves strong multi-shot generation after lightweight fine-tuning on our 60K dataset.
As demonstrated in~\Cref{fig:teaser}, 
this unified formulation naturally supports both text- and image-conditioned generation: the first shot can be initiated from text or text+image conditions, while subsequent shots are generated autoregressively as new captions are introduced.
The overall architecture is illustrated in~\Cref{fig:method}, while \Cref{sec:frame_selection} details the Frame Selection module, \Cref{sec:adaptive} introduces the Adaptive Conditioner, and \Cref{sec:training} describes our effective training strategy.

\vspace{0.5em}
\subsection{Frame Selection}
\label{sec:frame_selection}
A unique property of multi-shot videos is their \emph{unbounded spatio-temporal variance} across shots: adjacent shots are not necessarily contiguous in time or space~\citep{chasanis2008scene,kara2025shotadapter,guo2025long}.  
For example, Shot~1 may depict the protagonist, Shot~2 a secondary character, and Shot~3 the protagonist again.  
When generating Shot~3, the model should primarily reference Shot~1 to ensure subject consistency, while Shot~2 is less relevant.  
Motivated by such variable cross-shot relevance, we introduce a \emph{Frame Selection} module that selects semantically relevant frames from prior shots, ensuring appropriate visual context for consistent generation.

\noindent\textbf{Historical memory.}
When generating the $i$-th shot, we first encode all preceding shots $\{S_j\}_{j=1}^{i-1}$ by the 3D VAE encoder $\mathcal{E}$ to obtain latent features.  
For simplicity, we refer to these latent frames directly as “frames” in the following.
These encoded frames are concatenated into a global memory:
\begin{equation}
\label{eq:memory}
\mathbf{M}=\mathcal{F}_{\mathrm{concat}}\Big(\big\{\mathbf{z}_j^{(1)},\dots,\mathbf{z}_j^{(\frac{K}{f_t})}\big\}_{j=1}^{i-1}\Big)\in\mathbb{R}^{F\times N_s\times D_v},
\end{equation}
where $\mathbf{z}_j^{(\tau)}$ is the $\tau$-th frame feature of shot $S_j$,  
$F=(i-1)\tfrac{K}{f_t}$ is the total number of frames across preceding $i-1$ shots, and  
$N_s=\tfrac{H}{f_s}\times\tfrac{W}{f_s}$ is the number of tokens per frame.  
$\mathcal{F}_{\mathrm{concat}}$ concatenates features along the temporal axis, forming a unified visual memory of historical context.

\noindent\textbf{Querying with caption and memory.}
To identify relevant historical frames, we introduce $m$ learnable query tokens $\mathbf{Q}\in\mathbb{R}^{m\times D}$, where $D$ is the channel dimension of the model.  
These queries first attend to the current caption tokens to capture the semantic intent of the current shot:
\begin{equation}
\label{eq:q_text_attn}
\mathbf{Q}'=\mathcal{F}_{\mathrm{attn}}\!\left(\mathbf{Q},\,\phi_T(\mathbf{t}_i),\,\phi_T(\mathbf{t}_i)\right),
\end{equation}
where $\phi_T:\mathbb{R}^{D_t}\!\rightarrow\!\mathbb{R}^{D}$ projects text features into the latent space of the model, and $\mathcal{F}_{\mathrm{attn}}$ denotes the attention operation~\citep{vaswani2017attention} (first argument: queries; remaining: keys/values).
The updated queries $\mathbf{Q}'$ then attend to the visual memory to extract corresponding visual cues:
\begin{equation}
\label{eq:q_mem_attn}
\mathbf{M}_1=\phi_V(\mathbf{M}),\quad
\mathbf{Q}''=\mathcal{F}_{\mathrm{attn}}\!\left(\mathbf{Q}',\,\mathbf{M}_1,\,\mathbf{M}_1\right),
\end{equation}
where $\phi_V:\mathbb{R}^{D_v}\!\rightarrow\!\mathbb{R}^{D}$ is a convolutional projector that reduces the number of spatial tokens $N_s$ for efficiency.

\noindent\textbf{Scoring and selection.}
With the semantically and visually enriched queries, we compute frame-wise relevance scores to the current shot via query–memory interactions:
\begin{equation}
\label{eq:score}
\mathbf{s}=\phi_P(\mathbf{M}_1)\,\mathbf{Q}''^{\!\top}\in\mathbb{R}^{F\times m},\qquad
\mathbf{S}=\mathcal{F}_{\mathrm{mean}}(\mathbf{s})\in\mathbb{R}^{F},
\end{equation}
where $\phi_P$ projects each frame into a global embedding, and $\mathcal{F}_{\mathrm{mean}}$ averages scores across queries. 
To assist the learning of $\mathbf{S}$, we construct pseudo-labels indicating frame relevance to the current shot using DINOv2~\citep{oquab2023dinov2} and CLIP~\citep{radford2021learning} (more details in Appendix).
Next, the top-$K_\mathrm{sel}$ frames are selected from $\mathbf{M}$ based on $\mathbf{S}$ via the operation $\mathcal{F}_{\mathrm{TopK}}$:
\begin{equation}
\label{eq:mem}
\widehat{\mathbf{M}}=\mathcal{F}_{\mathrm{TopK}}(\mathbf{M},\,\mathbf{S},\,K_\mathrm{sel})\in\mathbb{R}^{K_\mathrm{sel}\times D}.
\end{equation}
The resulting memory $\widehat{\mathbf{M}}$ forms a semantically-relevant historical context, which is then passed to the \textit{Adaptive Conditioner} (\Cref{sec:adaptive}) for effective conditioning.

\subsection{Adaptive Conditioner}
\label{sec:adaptive}
Although the context memory $\widehat{\mathbf{M}}$ contains rich semantic cues, directly using all its tokens as conditioning signals incurs substantial computational overhead.  
To address this, we design an \emph{Adaptive Conditioner} which produces a compact set of context tokens with effective conditioning injection, balancing efficiency and informativeness.

\begin{figure}[t!]
    \hspace{2.3cm}
    \includegraphics[width=0.68\linewidth]{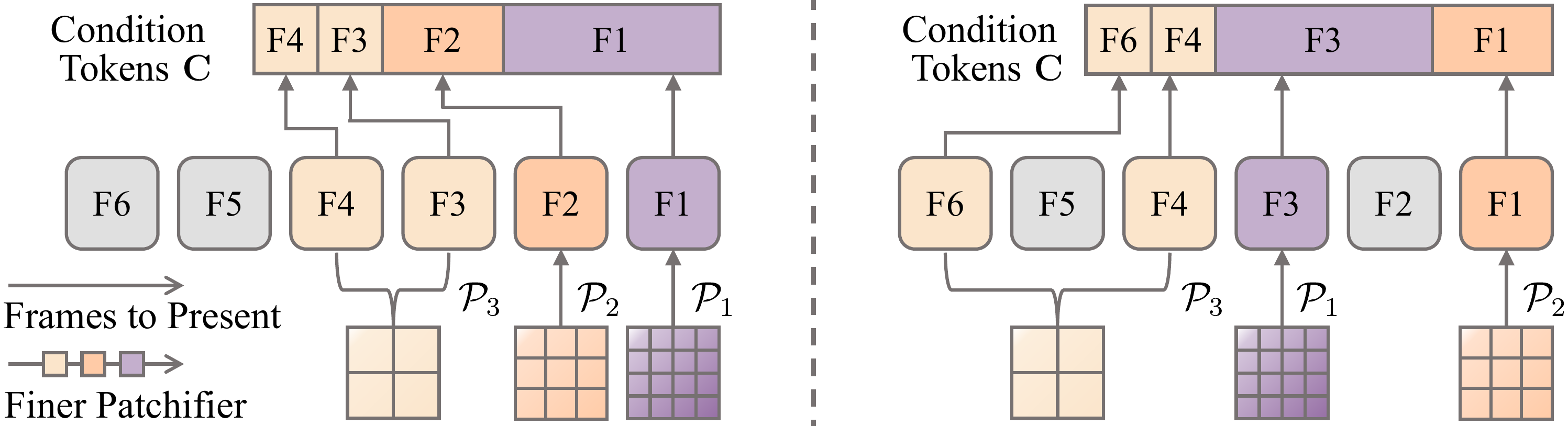}
    \caption{
    \textbf{Patchification Comparison.}
    \textit{Left:} Prior~\textit{fixed temporal schemes} typically consider the most recent block of contiguous frames and assign patchifiers by temporal order (\eg, the finest patchifier for the latest frame).
    \textit{Right:} Our~\textit{adaptive scheme} selects non-contiguous frames and allocates patchifiers based on content importance (\ie, finest patchifier for the most-important frame).
    }
    \label{fig:patchify}
\end{figure}

\noindent\textbf{Adaptive patchification.}
We define a set of patchifiers $\{\mathcal{P}_\ell\}_{\ell=1}^{L_p}$, each with a distinct kernel size.
Based on the relevance scores $\mathbf{S}$ from~\Cref{eq:score}, we divide the indices of $\widehat{\mathbf{M}}$ into $L_p$ disjoint subsets $\{\mathcal{I}_\ell\}_{\ell=1}^{L_p}$, assigning highly relevant frames to finer patchifiers with lower compression.
As in~\Cref{fig:patchify}, unlike prior fixed, temporal-based assignments~\citep{gu2025long,zhang2025packing}, our scheme adaptively allocates patchifiers to semantically relevant, non-adjacent frames, enabling content-driven rather than temporal-driven conditioning.
Each patchifier then transforms its assigned frames into context tokens:
\begin{equation}
\label{eq:cond_tokens}
\mathbf{C}_\ell = \mathcal{P}_\ell\!\big(\widehat{\mathbf{M}}_{\mathcal{I}_\ell}\big), \qquad
\mathbf{C} = \mathcal{F}_{\mathrm{concat}}\!\big(\{\mathbf{C}_\ell\}_{\ell=1}^{L_p}\big),
\end{equation}
where $\mathbf{C}_\ell \in \mathbb{R}^{N_\ell \times D}$ and $\mathbf{C} \in \mathbb{R}^{N_c \times D}$,  
with $N_\ell$ denoting the number of tokens produced by $\mathcal{P}_\ell$ and $N_c = \sum_{\ell=1}^{L_p} N_\ell$ the total number of context tokens.

\noindent\textbf{Condition injection.}
Let $\mathbf{N}\in\mathbb{R}^{N_n\times D}$ denote the noise tokens of the current shot in the diffusion process~\citep{ho2020denoising}.  
We concatenate the context tokens $\mathbf{C}$ with $\mathbf{N}$ along the token dimension to form the DiT~\citep{peebles2023scalable} input:
\begin{equation}
\label{eq:concat}
\mathbf{X} = \mathcal{F}_{\mathrm{concat}}([\mathbf{N},\,\mathbf{C}]) \in \mathbb{R}^{(N_n + N_c) \times D}.
\end{equation}
This simple yet effective injection scheme enables joint attention between noisy and context tokens, facilitating rich interactions.  
By adjusting the patchifiers $\{\mathcal{P}_\ell\}_{\ell=1}^{L_p}$ to balance compression and retention, the additional computation remains minimal.  
Overall, the Adaptive Conditioner provides efficient, relevance-aware conditioning that achieves compactness without sacrificing context expressiveness.

\subsection{Training Strategy}
\label{sec:training}

We train the model \emph{jointly and end-to-end} by predicting the final shot in each sequence conditioned on its preceding shots.  
To ensure effective optimization and enhance narrative consistency, we introduce the following training strategies.  
More details are in Appendix.

\noindent\textbf{Shot inflation.}
The dataset introduced in \Cref{sec:data_process} contains videos with varying numbers of shots, primarily two-shot sequences with fewer three-shot ones, which destabilizes optimization when trained jointly.  
Therefore, we \emph{inflate} two-shot sequences into three-shot ones by either (i) inserting a shot sampled from another video or (ii) augmenting the first shot (\eg, spatial or color transformations).
This process yields mixed real and synthetic triplets, enabling uniform three-shot training.

\noindent\textbf{Decoupled conditioning.}
Early in training, the Frame Selection module is randomly initialized and may select suboptimal frames, complicating optimization.  
We introduce a \emph{two-stage curriculum}.  
During warm-up, we train on synthetic three-shot sequences and uniformly sample conditioning frames from the real shot, effectively decoupling conditioning from the selector’s outputs.
Afterward, we switch to full \emph{selector-driven conditioning}, where selected frames directly guide generation.  
This progressive coupling stabilizes convergence and enhances narrative coherence.

\section{Experiments}
\subsection{Experimental Setup}
\vspace{0.5em}
\noindent\textbf{Implementation Details.}
Our model builds upon the pretrained I2V model Wan2.1~\citep{wan2025wan}.
We optimize using AdamW with a learning rate of 0.0005 and weight decay of 0.01.
The entire model is fine-tuned for one epoch on 128 NVIDIA A100 GPUs using our curated multi-shot dataset.  
All videos are center-cropped to $480{\times}832$ while preserving aspect ratio.
To comprehensively evaluate MSV, we construct dedicated benchmarks for both T2MSV and I2MSV settings, covering diverse human-centric narratives with complex cross-shot dynamics such as reappearance and composition.
Further details are in Appendix.

\begin{table*}[tb]
  \footnotesize
  \centering
  \renewcommand{\arraystretch}{1.4}%
  \resizebox{\textwidth}{!}{%
  \begin{tabular}{
    p{3.7cm}
    *{9}{>{\centering\arraybackslash}m{1.0cm}}
  }
    \toprule
    \multirow{2}{*}{\textbf{Method}}
      & \multicolumn{3}{c}{\textbf{Inter-shot Coherence}}
      & \multirow{2}{*}{\parbox{1.05cm}{\centering \textbf{Semantic\\Align.}\textuparrow}}
      & \multicolumn{3}{c}{\textbf{Intra-shot Coherence}}
      & \multirow{2}{*}{\parbox{1.05cm}{\centering \textbf{Aesthetic\\Quality}\textuparrow}}
      & \multirow{2}{*}{\parbox{1.05cm}{\centering \textbf{Dynamic\\Degree}\textuparrow}} \\
    \cline{2-4} \cline{6-8}
      & \rule{0pt}{1.25em}\textbf{Character\textuparrow}
      & \rule{0pt}{1.25em}\textbf{Env.\textuparrow}
      & \rule{0pt}{1.25em}\textbf{Avg.\textuparrow}
      &
      & \rule{0pt}{1.25em}\textbf{Subject\textuparrow}
      & \rule{0pt}{1.25em}\textbf{BG.\textuparrow}
      & \rule{0pt}{1.25em}\textbf{Avg.\textuparrow}
      &
      & \\
    \midrule
    \noalign{\vspace{-0.7ex}}
    \rowcolor{gray!10}
    \multicolumn{10}{l}{\textit{Text-to-multi-shot (T2MSV)}} \\
    Flux + LTX-Video           & 0.5316 & 0.5456 & 0.5386 & 0.1837 & 0.8841 & 0.8957 & 0.8899 & 0.5070 & 0.3746 \\
    Flux + Wan2.1              & 0.5454 & 0.5598 & 0.5526 & 0.1915 & 0.9225 & 0.9353 & 0.9289 & 0.5572 & \underline{0.4492} \\
    Mask$^2$DiT                & 0.5472 & 0.5419 & 0.5446 & \underline{0.2253} & 0.9024 & 0.9150 & 0.9087 & 0.5235 & 0.4247 \\
    StoryDiff. + LTX-Video & 0.5468 & 0.5397 & 0.5433 & 0.2165 & 0.9036 & 0.9125 & 0.9081 & 0.5418 & 0.3694 \\
    StoryDiff. + Wan2.1          & \underline{0.5633} & \underline{0.5681} & \underline{0.5657} & 0.2217 & \underline{0.9286} & \underline{0.9357} & \underline{0.9322} & \underline{0.5703} & 0.4231 \\
    \rowcolor{row} \ourmodel~(Ours)           & \textbf{0.5874} & \textbf{0.5752} & \textbf{0.5813} & \textbf{0.2389} & \textbf{0.9364} & \textbf{0.9410} & \textbf{0.9387} & \textbf{0.5731} & \textbf{0.4698} \\
    \midrule
    \noalign{\vspace{-0.7ex}}
    \rowcolor{gray!10}
    \multicolumn{10}{l}{\textit{Image-to-multi-shot (I2MSV)}} \\
    Mask$^2$DiT                & \underline{0.5452} & 0.5446 & 0.5449 & \underline{0.2270} & 0.9056 & 0.9073 & 0.9065 & 0.5218 & 0.4256 \\
    Flux + LTX-Video           & 0.5336 & 0.5469 & 0.5403 & 0.1846 & 0.8803 & 0.8930 & 0.8867 & 0.5114 & 0.3790 \\
    Flux + Wan2.1              & 0.5419 & \underline{0.5547} & \underline{0.5483} & 0.1897 & \underline{0.9186} & \underline{0.9310} & \underline{0.9248} & \underline{0.5531} & \underline{0.4544} \\
    \rowcolor{row} \ourmodel~(Ours)           & \textbf{0.5851} & \textbf{0.5716} & \textbf{0.5784} & \textbf{0.2354} & \textbf{0.9327} & \textbf{0.9389} & \textbf{0.9358} & \textbf{0.5704} & \textbf{0.4673} \\
    \bottomrule
  \end{tabular}%
  }
  \caption{
  \textbf{Quantitative results under text-to-multi-shot (T2MSV) and image-to-multi-shot (I2MSV) settings.}
The best and runner-up results are in \textbf{bold} and \underline{underlined}, respectively.
In both text- and image-conditioned settings, our model consistently outperforms all baselines on shot-level quality and narrative consistency, demonstrating superior multi-shot generation capabilities.
``Env.'' denotes environment consistency, ``BG.'' denotes background consistency, and ``Avg.'' indicates the average of the corresponding metrics.
  }
  \label{tab:multishot_results}
\end{table*}

\begin{figure*}[t!] 
    \centering
    \includegraphics[width=\textwidth]{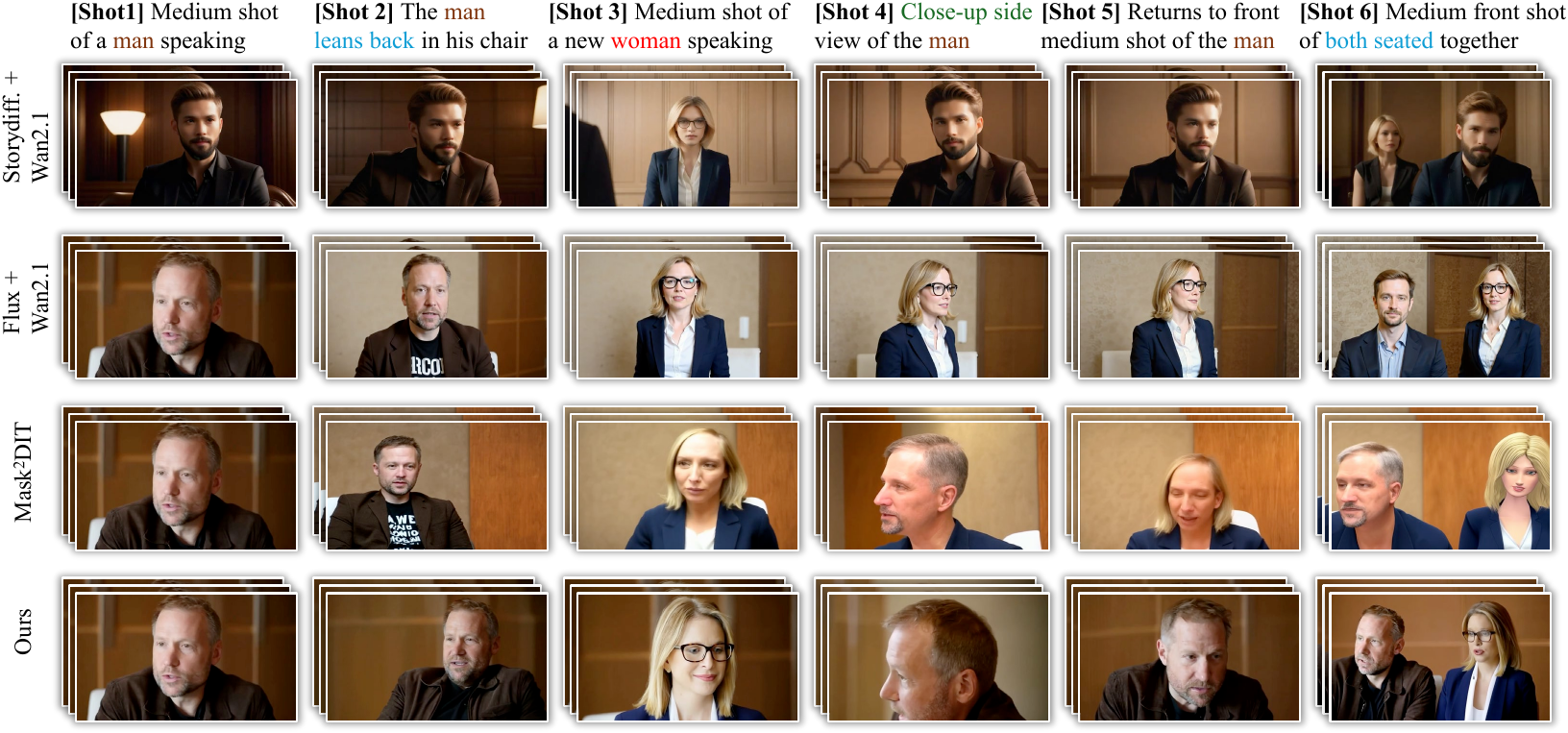} 
    \caption{
    \textbf{Qualitative results.}
For a fair comparison, the given multi-shot generations share the same first shot (generated by \textit{Wan2.2}) as the initial condition, except for \textit{StoryDiff.+Wan2.1}, which does not rely on visual conditioning. The baseline methods fail to maintain narrative consistency across shots, struggling with prompt adherence, reappearance, and compositional scenes, whereas \ourmodel~(ours) faithfully follows shot-level captions and produces coherent shots. A representative segment of each prompt is given with the corresponding shot.
} 
    \label{fig:comparison} 
\end{figure*}

\begin{table*}[tb]
  \footnotesize
  \centering
  \begin{subtable}[t]{0.29\textwidth}
    \centering
    \setlength{\tabcolsep}{5.5pt}
    \renewcommand{\arraystretch}{1.1}
    \begin{tabular}{ccccc}
      \toprule
      \textbf{AC} & \textbf{FS} & \textbf{C-Cons}\,$\uparrow$ & \textbf{E-Cons}\,$\uparrow$ & \textbf{S-Align}\,$\uparrow$ \\
      \midrule
                  &             & 0.5153 & 0.5112 & 0.1814 \\
      \checkmark  &             & 0.5465 & 0.5597 & 0.2172 \\
                  & \checkmark  & 0.5526 & 0.5710 & 0.2238 \\
      \rowcolor{row} \checkmark  & \checkmark  & \textbf{0.5874} & \textbf{0.5752} & \textbf{0.2389} \\
      \bottomrule
    \end{tabular}
    \subcaption{Impact of model design.}
    \label{tab:ab_design}
  \end{subtable}
  \hspace{0.072\textwidth}%
  \begin{subtable}[t]{0.29\textwidth}
    \centering
    \setlength{\tabcolsep}{5.5pt}
    \renewcommand{\arraystretch}{1.37}
    \begin{tabular}{ccccc}
      \toprule
      \textbf{SI} & \textbf{DC} & \textbf{C-Cons}\,$\uparrow$ & \textbf{E-Cons}\,$\uparrow$ & \textbf{S-Align}\,$\uparrow$ \\
      \midrule
                  &             & 0.5514 & 0.5615 & 0.2207 \\
      \checkmark  &             & 0.5649 & \textbf{0.5790} & 0.2263 \\
      \rowcolor{row} \checkmark  & \checkmark  & \textbf{0.5874} & 0.5752 & \textbf{0.2389} \\
      \bottomrule
    \end{tabular}
    \subcaption{Impact of training strategies.}
    \label{tab:ab_train}
  \end{subtable}
  \hspace{0.057\textwidth}%
  \begin{subtable}[t]{0.28\textwidth}
    \centering
    \setlength{\tabcolsep}{5.5pt}
    \renewcommand{\arraystretch}{1.37}
    \begin{tabular}{ccc}
      \toprule
      \textbf{Ctx len} & \textbf{C-Cons}\,$\uparrow$ & \textbf{E-Cons}\,$\uparrow$ \\
      \midrule
      1-frame & 0.5874 & 0.5752 \\
      2-frame & 0.5890 & 0.5795 \\
      \rowcolor{row} 3-frame & \textbf{0.5926} & \textbf{0.5863} \\
      \bottomrule
    \end{tabular}
    \subcaption{Impact of \#context tokens.}
    \label{tab:ab_context}
  \end{subtable}
  \caption{\textbf{Ablation study.}
(a) Model design: both Adaptive Conditioner (AC) and Frame Selection (FS) are crucial for multi-shot generation quality.
(b) Training strategies: Shot Inflation (SI) and Decoupled Conditioning (DC) improve narrative learning.
(c) Context token length (Ctx len): with one latent frame as the unit of context token budget, a single-frame equivalent performs strongly, and more tokens further improve.
C-Cons and E-Cons denote character/environment consistency, with S-Align as semantic alignment.
Best results are in \textbf{bold}.}
\end{table*}

\begin{figure*}[t!]
    \centering

    \begin{minipage}[c]{0.48\linewidth}
        \centering
        \includegraphics[width=\linewidth]{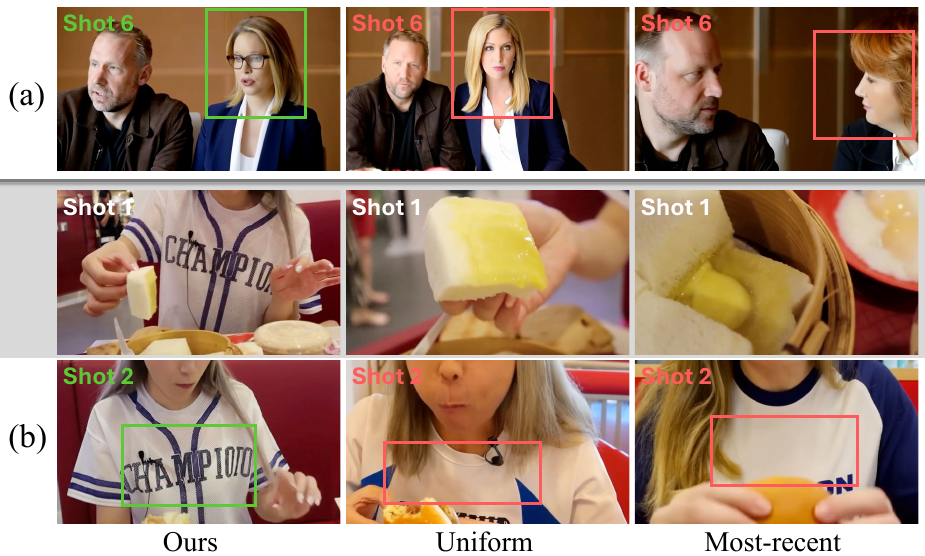}
        \captionof{figure}{\textbf{Qualitative comparison of frame selection strategies.} 
(a) The sixth-shot generation using the five preceding shots as context (shown in the last row of \Cref{fig:comparison}). (b) Fine-grained cases where the first shot involves dynamic motion. In each, the first gray row shows a few sampled frames (first, middle, and last) depicting the motion range in the first shot. The subsequent row shows the generated next shot from each strategy. Both baselines fail to maintain visual coherence, whereas our method identifies semantically relevant frames and produces consistent shots.
}
        \label{fig:vis1}
    \end{minipage}
    \hfill
    \begin{minipage}[c]{0.48\linewidth}
        \centering
        \includegraphics[width=\linewidth]{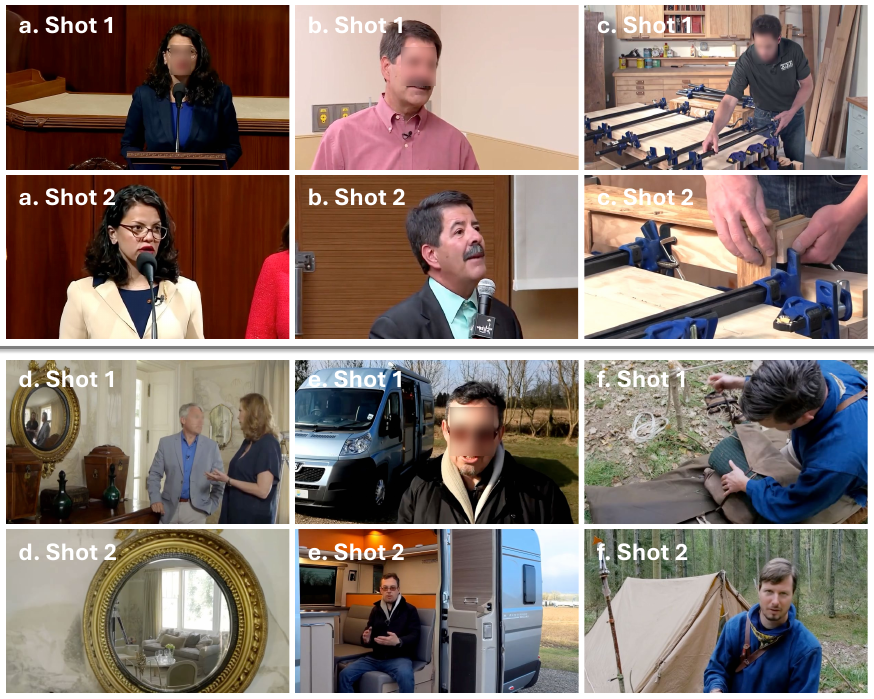}
        \captionof{figure}{\textbf{Advanced narrative modeling in \ourmodel.}
(a-b)~\textit{Appearance changes:} the character’s identity remains consistent under appearance variations.
(c–d)~\textit{Zoom-in effects:} the model accurately localizes intended regions and preserves fine details when zooming in.
(e–f)~\textit{Human–object interactions:} the model correctly continues event progression, maintaining coherent relationships between humans and surrounding objects across shots.
}
        \label{fig:vis2}
    \end{minipage}

\end{figure*}

\noindent\textbf{Baselines.}
We compare \ourmodel~with three strong MSV paradigms, using a vision-language model~\citep{llama4,bai2025qwen2,yuan2025tarsier2} to convert shot-level captions into method-specific prompts:

\begin{enumerate}
\item \textit{Fixed-window attention} extends attention to multiple shots within a fixed temporal window.  
We employ the public Mask$^2$DiT~\citep{qi2025mask} as a representative baseline.

\item \textit{Keyframe conditioning} first generates a keyframe per shot, which is then expanded into a shot by an I2V model.
We use StoryDiffusion~\citep{zhou2024storydiffusion} for keyframes and Wan~2.1~\citep{wan2025wan}~/~LTX-Video~\citep{hacohen2024ltx} for I2V synthesis.

\item \textit{Edit-and-extend} treats MSV as next-shot generation, transferring the last frame of the previous shot via an I2I model before I2V synthesis.
We use FLUX~\citep{flux} as the I2I model and Wan2.1~/~LTX-Video for I2V generation.
\end{enumerate}

\subsection{Main Results}
\noindent\textbf{Quantitative Evaluation.}
We evaluate from two perspectives: \textit{shot-level quality} and \textit{narrative consistency}.
Shot-level quality follows single-shot metrics~\citep{huang2024vbench}, including \textit{subject consistency}, \textit{background consistency}, \textit{aesthetic quality}, and \textit{dynamic degree}.
For narrative consistency, we design metrics following prior studies~\citep{kara2025shotadapter}:
\begin{enumerate}
\item \textit{Character Consistency} computes DINOv2~\citep{oquab2023dinov2} similarity between YOLO~\citep{ultralytics2021yolov5} segmented persons across shots annotated as containing the same character.  
\item \textit{Environment Consistency} measures DINOv2 similarity between segmented background regions across shots annotated with matched environments.  
\item \textit{Semantic Alignment} quantifies the alignment between each generated shot and its caption using ViCLIP~\citep{wanginternvid}.  
\end{enumerate}
We group subject/background consistency as \textit{intra-shot coherence} and character/environment consistency as \textit{inter-shot coherence}.
As shown in~\Cref{tab:multishot_results}, for T2MSV, our model significantly surpasses all baselines across inter-shot coherence and semantic alignment metrics, demonstrating superior narrative consistency.  
It also achieves better shot-level performance with improved motion control and intra-shot fidelity.
In I2MSV, we omit keyframe baselines since they cannot directly accept image inputs.  
Our model again attains state-of-the-art results across both shot-level and narrative metrics, further confirming its effectiveness.

\vspace{0.5em}
\noindent\textbf{Qualitative Results.}
\Cref{fig:comparison} shows qualitative comparisons under complex narratives.
All baselines struggle to follow shot-level prompts and maintain cross-shot coherence.
For instance, in Shot 4, both StoryDiff.~\citep{zhou2024storydiffusion} + Wan2.1~\citep{wan2025wan} and Flux~\citep{flux} + Wan2.1 fail to adjust the viewpoint accordingly, while Mask$^2$DiT~\citep{qi2025mask} and Flux+Wan2.1 generate the wrong character in Shot 5.
When Shot 3 introduces a new character (the woman), the baselines further lose coherence when the protagonist (the man) reappears in Shots 4–6.
Moreover, in the final compositional shot requiring both characters to appear together, they all collapse in maintaining character identity, revealing limited memory adaptation ability.
In contrast, \ourmodel~maintains subject and environment consistency across reappearances and compositions, while faithfully adhering to evolving narratives, highlighting the superiority of our adaptive memory for coherent narrative generation.

\subsection{Ablation Study}
\noindent\textbf{Impact of model design.}
\Cref{tab:ab_design} analyzes each design under the same context-token budget, equal to the number of tokens in one latent frame.
The baseline relies only on the last frame, showing the weakest performance due to missing historical context.
Adding the \textit{Adaptive Conditioner} extends contextual range, while the \textit{Frame Selection} module enhances adaptation by conditioning on the most relevant frame.
Combining both yields the best results, confirming their complementary roles in cross-shot context modeling.

\noindent\textbf{Frame selection.}
\Cref{fig:vis1} compares our automatic frame selection with uniform and most-recent sampling strategies.
Each variant is trained with its own sampling scheme under the same context budget.
In \Cref{fig:vis1}(a), the first five shots from the last row of \Cref{fig:comparison} serve as history for predicting the sixth shot.
\Cref{fig:vis1}(b) shows a more challenging fine-grained case where the first shot contains large camera motion (see the $2^{nd}$ gray row), requiring precise frame selection to preserve continuity.
Across both cases, our method maintains contextual consistency, while both baselines fail, highlighting the effectiveness of our frame selection module.

\noindent\textbf{Impact of training strategies.}
\Cref{tab:ab_train} evaluates proposed training strategies.
Training on mixed two- and three-shot videos (baseline) hinders narrative learning due to imbalanced context sequences, whereas \textit{shot inflation} enables unified three-shot training with richer temporal context and improves generation ability.
Adding the \textit{decoupled conditioning} curriculum further stabilizes early optimization and strengthens narrative coherence.

\noindent\textbf{Context tokens.}
We vary the number of context tokens in the \textit{Adaptive Conditioner} by adjusting its patchifiers.
We define one latent frame as the unit of context token amount to allow direct comparison with the number of original noise tokens (21-frame).
As shown in \Cref{tab:ab_context}, even a single latent-frame equivalent of context tokens yields strong inter-shot coherence, and larger budgets further enhance performance.
This confirms the efficiency of our compact adaptive memory in modeling cross-shot dynamics.
By default, we use one latent-frame equivalent of context tokens.

\subsection{Advanced Narrative Modeling}
Real-world narratives exhibit complex cross-shot dependencies.
\ourmodel~captures these dynamics through adaptive memory, showing advanced generation ability with a global narrative understanding beyond surface-level visual continuity.
We analyze its advanced narrative modeling from three perspectives.

\noindent\textbf{Appearance changes.}
Beyond typical cross-shot variations (\eg, viewpoint), maintaining character consistency under appearance changes is especially challenging.
As shown in \Cref{fig:vis2}(a–b), \ourmodel~preserves consistent facial features while adapting clothing and environments as prompted, showcasing robust narrative coherence under complex visual changes. We refer readers to \projectpage for better visualizations.

\noindent\textbf{Zoom-in effects.}  
Transitions from wide shots to non-salient close-ups demand spatial reasoning to locate fine-grained targets while preserving fidelity.
In \Cref{fig:vis2}(c–d), \ourmodel~accurately identifies local regions and maintains detail for both static (\ie, mirror) and dynamic (\ie, hand) targets.
In \Cref{fig:teaser} ($3^{rd}$ example), small objects on the table, barely visible in Shot 2, are faithfully rendered in the zoomed-in Shot 3, yielding smooth narrative progression.

\noindent\textbf{Human-object interactions.}
Shot transitions often hinge on human–object dynamics and implicit intent.
In \Cref{fig:vis2}(e–f), scenes of a man engaging with a car or unfolding a tent lead to coherent next shots that depict the expected subsequent state.
These examples highlight \ourmodel’s ability to interpret human–object relations, enabling realistic and semantically grounded story development.

\section{Conclusion}
We presented \ourmodel, a novel framework for coherent multi-shot video generation via adaptive memory modeling.
By reformulating MSV as a next-shot generation problem, \ourmodel~leverages the strong visual conditioning capacity of pretrained I2V models, enabling scalable and autoregressive story synthesis.
The proposed Frame Selection module identifies semantically relevant frames across prior shots, while the Adaptive Conditioner performs importance-guided patchification with direct condition injection, jointly enabling global yet compact cross-shot context.
\ourmodel\ effectively handles complex narratives and achieves superior narrative coherence, offering valuable insights into adaptive memory modeling for immersive, story-driven video generation.

\section*{Acknowledgements}
We would like to thank Junlin Han (University of Oxford), Mingqiao Ye (EPFL), and Feng Qiao (WashU) for their constructive feedback on this project.
Zhaochong An and Serge Belongie are supported by funding from the Pioneer Centre for AI, DNRF grant number P1.

\bibliographystyle{assets/plainnat}
\bibliography{paper}

@String(IJCV = {Int. J. Comput. Vis.})

@String(CVPR= {IEEE Conf. Comput. Vis. Pattern Recog.})

@String(ICCV= {Int. Conf. Comput. Vis.})

@String(ECCV= {Eur. Conf. Comput. Vis.})

@String(ICLR = {Int. Conf. Learn. Represent.})

@String(IJCV  = {IJCV})

@String(CVPR  = {CVPR})

@String(ICCV  = {ICCV})

@String(ECCV  = {ECCV})

@String(ICLR  = {ICLR})

@article{zhou2024storydiffusion,
  title={StoryDiffusion: Consistent Self-Attention for Long-Range Image and Video Generation},
  author={Zhou, Yupeng and Zhou, Daquan and Cheng, Ming-Ming and Feng, Jiashi and Hou, Qibin},
  journal={NeurIPS 2024},
  year={2024}
}

@miscf{llama4,
    title={The Llama 4 herd: The beginning of a new era of natively multimodal AI innovation},
    author={Team Meta},
    howpublished={\url{https://ai.meta.com/blog/llama-4-multimodal-intelligence/}},
    month={April},
    year={2025}
}

@article{yuan2025tarsier2,
  title={Tarsier2: Advancing large vision-language models from detailed video description to comprehensive video understanding},
  author={Yuan, Liping and Wang, Jiawei and Sun, Haomiao and Zhang, Yuchen and Lin, Yuan},
  journal={arXiv preprint},
  year={2025}
}

@inproceedings{wang2025cinemaster,
  title={Cinemaster: A {3D}-aware and controllable framework for cinematic text-to-video generation},
  author={Wang, Qinghe and Luo, Yawen and Shi, Xiaoyu and Jia, Xu and Lu, Huchuan and Xue, Tianfan and Wang, Xintao and Wan, Pengfei and Zhang, Di and Gai, Kun},
  booktitle={SIGGRAPH},
  pages={1--10},
  year={2025}
}

@article{wei2025mocha,
  title={Mocha: Towards movie-grade talking character synthesis},
  author={Wei, Cong and Sun, Bo and Ma, Haoyu and Hou, Ji and Juefei-Xu, Felix and He, Zecheng and Dai, Xiaoliang and Zhang, Luxin and Li, Kunpeng and Hou, Tingbo and others},
  journal={arXiv preprint},
  year={2025}
}

@article{zhang2025shouldershot,
  title={ShoulderShot: Generating Over-the-Shoulder Dialogue Videos},
  author={Zhang, Yuang and Cheng, Junqi and Zhao, Haoyu and Gu, Jiaxi and Zou, Fangyuan and Lu, Zenghui and Shu, Peng},
  journal={arXiv preprint},
  year={2025}
}

@inproceedings{soucek2024transnet,
  title={Transnet v2: An effective deep network architecture for fast shot transition detection},
  author={Soucek, Tom{\'a}s and Lokoc, Jakub},
  booktitle={ACM MM},
  pages={11218--11221},
  year={2024}
}

@misc{ultralytics2021yolov5,
  author       = {Ultralytics},
  title        = {{YOLOv5}: A state-of-the-art real-time object detection system},
  howpublished = {\url{https://docs.ultralytics.com}},
  year         = {2021},
  note         = {Accessed: 2024-11-04}
}

@inproceedings{wanginternvid,
  title={InternVid: A Large-scale Video-Text Dataset for Multimodal Understanding and Generation},
  author={Wang, Yi and He, Yinan and Li, Yizhuo and Li, Kunchang and Yu, Jiashuo and Ma, Xin and Li, Xinhao and Chen, Guo and Chen, Xinyuan and Wang, Yaohui and others},
  booktitle={ICLR},
  year={2024}
}

@article{gu2025long,
  title={Long-context autoregressive video modeling with next-frame prediction},
  author={Gu, Yuchao and Mao, Weijia and Shou, Mike Zheng},
  journal={arXiv preprint},
  year={2025}
}

@article{jia2025moga,
  title={MoGA: Mixture-of-Groups Attention for End-to-End Long Video Generation},
  author={Jia, Weinan and Lu, Yuning and Huang, Mengqi and Wang, Hualiang and Huang, Binyuan and Chen, Nan and Liu, Mu and Jiang, Jidong and Mao, Zhendong},
  journal={arXiv preprint},
  year={2025}
}

@article{meng2025holocine,
  title={HoloCine: Holistic Generation of Cinematic Multi-Shot Long Video Narratives},
  author={Meng, Yihao and Ouyang, Hao and Yu, Yue and Wang, Qiuyu and Wang, Wen and Cheng, Ka Leong and Wang, Hanlin and Li, Yixuan and Chen, Cheng and Zeng, Yanhong and others},
  journal={arXiv preprint},
  year={2025}
}

@article{song2025lavieid,
  title={LaVieID: Local Autoregressive Diffusion Transformers for Identity-Preserving Video Creation},
  author={Song, Wenhui and Li, Hanhui and Huang, Jiehui and Hu, Panwen and Cheng, Yuhao and Chen, Long and Yan, Yiqiang and Liang, Xiaodan},
  journal={arXiv preprint},
  year={2025}
}

@article{ho2020denoising,
  title={Denoising diffusion probabilistic models},
  author={Ho, Jonathan and Jain, Ajay and Abbeel, Pieter},
  journal={NeurIPS},
  volume={33},
  pages={6840--6851},
  year={2020}
}

@article{vaswani2017attention,
  title={Attention is all you need},
  author={Vaswani, Ashish and Shazeer, Noam and Parmar, Niki and Uszkoreit, Jakob and Jones, Llion and Gomez, Aidan N and Kaiser, {\L}ukasz and Polosukhin, Illia},
  journal={NeurIPS},
  volume={30},
  year={2017}
}

@misc{kling,
  author={{Kuaishou}},
  title={Kling Video Model},
  year={2024},
  howpublished={\url{https://kling.kuaishou.com/en}},
}

@misc{flux,
  author={{Black Forest Labs}},
  title={FLUX},
  year={2024},
  howpublished={\url{https://github.com/black-forest-labs/flux}},
}

@article{xiao2025captain,
  title={Captain cinema: Towards short movie generation},
  author={Xiao, Junfei and Yang, Ceyuan and Zhang, Lvmin and Cai, Shengqu and Zhao, Yang and Guo, Yuwei and Wetzstein, Gordon and Agrawala, Maneesh and Yuille, Alan and Jiang, Lu},
  journal={arXiv preprint},
  year={2025}
}

@article{yu2025context,
  title={Context as memory: Scene-consistent interactive long video generation with memory retrieval},
  author={Yu, Jiwen and Bai, Jianhong and Qin, Yiran and Liu, Quande and Wang, Xintao and Wan, Pengfei and Zhang, Di and Liu, Xihui},
  journal={arXiv preprint},
  year={2025}
}

@article{he2025cut2next,
  title={Cut2Next: Generating Next Shot via In-Context Tuning},
  author={He, Jingwen and Liu, Hongbo and Li, Jiajun and Huang, Ziqi and Qiao, Yu and Ouyang, Wanli and Liu, Ziwei},
  journal={arXiv preprint},
  year={2025}
}

@article{gao2025seedance,
  title={Seedance 1.0: Exploring the Boundaries of Video Generation Models},
  author={Gao, Yu and Guo, Haoyuan and Hoang, Tuyen and Huang, Weilin and Jiang, Lu and Kong, Fangyuan and Li, Huixia and Li, Jiashi and Li, Liang and Li, Xiaojie and others},
  journal={arXiv preprint},
  year={2025}
}

@article{jiang2025lovic,
  title={Lovic: Efficient long video generation with context compression},
  author={Jiang, Jiaxiu and Li, Wenbo and Ren, Jingjing and Qiu, Yuping and Guo, Yong and Xu, Xiaogang and Wu, Han and Zuo, Wangmeng},
  journal={arXiv preprint},
  year={2025}
}

@article{cai2025mixture,
  title={Mixture of contexts for long video generation},
  author={Cai, Shengqu and Yang, Ceyuan and Zhang, Lvmin and Guo, Yuwei and Xiao, Junfei and Yang, Ziyan and Xu, Yinghao and Yang, Zhenheng and Yuille, Alan and Guibas, Leonidas and others},
  journal={arXiv preprint},
  year={2025}
}

@article{tschannen2025siglip,
  title={Siglip 2: Multilingual vision-language encoders with improved semantic understanding, localization, and dense features},
  author={Tschannen, Michael and Gritsenko, Alexey and Wang, Xiao and Naeem, Muhammad Ferjad and Alabdulmohsin, Ibrahim and Parthasarathy, Nikhil and Evans, Talfan and Beyer, Lucas and Xia, Ye and Mustafa, Basil and others},
  journal={arXiv preprint},
  year={2025}
}

@inproceedings{radford2021learning,
  title={Learning transferable visual models from natural language supervision},
  author={Radford, Alec and Kim, Jong Wook and Hallacy, Chris and Ramesh, Aditya and Goh, Gabriel and Agarwal, Sandhini and Sastry, Girish and Askell, Amanda and Mishkin, Pamela and Clark, Jack and others},
  booktitle={ICML},
  pages={8748--8763},
  year={2021},
}

@article{oquab2023dinov2,
  title={Dinov2: Learning robust visual features without supervision},
  author={Oquab, Maxime and Darcet, Timoth{\'e}e and Moutakanni, Th{\'e}o and Vo, Huy and Szafraniec, Marc and Khalidov, Vasil and Fernandez, Pierre and Haziza, Daniel and Massa, Francisco and El-Nouby, Alaaeldin and others},
  journal={arXiv preprint},
  year={2023}
}

@article{qiu2025animeshooter,
  title={AnimeShooter: A Multi-Shot Animation Dataset for Reference-Guided Video Generation},
  author={Qiu, Lu and Li, Yizhuo and Ge, Yuying and Ge, Yixiao and Shan, Ying and Liu, Xihui},
  journal={arXiv preprint},
  year={2025}
}

@article{wu2025cinetrans,
  title={CineTrans: Learning to Generate Videos with Cinematic Transitions via Masked Diffusion Models},
  author={Wu, Xiaoxue and Gao, Bingjie and Qiao, Yu and Wang, Yaohui and Chen, Xinyuan},
  journal={arXiv preprint},
  year={2025}
}

@article{yang2024cogvideox,
  title={Cogvideox: Text-to-video diffusion models with an expert transformer},
  author={Yang, Zhuoyi and Teng, Jiayan and Zheng, Wendi and Ding, Ming and Huang, Shiyu and Xu, Jiazheng and Yang, Yuanming and Hong, Wenyi and Zhang, Xiaohan and Feng, Guanyu and others},
  journal={arXiv preprint},
  year={2024}
}

@inproceedings{chen2024videocrafter2,
  title={Videocrafter2: Overcoming data limitations for high-quality video diffusion models},
  author={Chen, Haoxin and Zhang, Yong and Cun, Xiaodong and Xia, Menghan and Wang, Xintao and Weng, Chao and Shan, Ying},
  booktitle={CVPR},
  pages={7310--7320},
  year={2024}
}

@article{brooks2024video,
  title={Video generation models as world simulators},
  author={Brooks, Tim and Peebles, Bill and Holmes, Connor and DePue, Will and Guo, Yufei and Jing, Li and Schnurr, David and Taylor, Joe and Luhman, Troy and Luhman, Eric and others},
  year={2024}
}

@inproceedings{peebles2023scalable,
  title={Scalable diffusion models with transformers},
  author={Peebles, William and Xie, Saining},
  booktitle={ICCV},
  pages={4195--4205},
  year={2023}
}

@article{polyak2024movie,
  title={Movie gen: A cast of media foundation models},
  author={Polyak, Adam and Zohar, Amit and Brown, Andrew and Tjandra, Andros and Sinha, Animesh and Lee, Ann and Vyas, Apoorv and Shi, Bowen and Ma, Chih-Yao and Chuang, Ching-Yao and others},
  journal={arXiv preprint},
  year={2024}
}

@inproceedings{hu2024animate,
  title={Animate anyone: Consistent and controllable image-to-video synthesis for character animation},
  author={Hu, Li},
  booktitle={CVPR},
  pages={8153--8163},
  year={2024}
}

@inproceedings{chen2023seine,
  title={Seine: Short-to-long video diffusion model for generative transition and prediction},
  author={Chen, Xinyuan and Wang, Yaohui and Zhang, Lingjun and Zhuang, Shaobin and Ma, Xin and Yu, Jiashuo and Wang, Yali and Lin, Dahua and Qiao, Yu and Liu, Ziwei},
  booktitle={ICLR},
  year={2023}
}

@inproceedings{zeng2024make,
  title={Make pixels dance: High-dynamic video generation},
  author={Zeng, Yan and Wei, Guoqiang and Zheng, Jiani and Zou, Jiaxin and Wei, Yang and Zhang, Yuchen and Li, Hang},
  booktitle={CVPR},
  pages={8850--8860},
  year={2024}
}

@inproceedings{xing2025motioncanvas,
  title={Motioncanvas: Cinematic shot design with controllable image-to-video generation},
  author={Xing, Jinbo and Mai, Long and Ham, Cusuh and Huang, Jiahui and Mahapatra, Aniruddha and Fu, Chi-Wing and Wong, Tien-Tsin and Liu, Feng},
  booktitle={Proceedings of the Special Interest Group on Computer Graphics and Interactive Techniques Conference Conference Papers},
  pages={1--11},
  year={2025}
}

@article{bai2025qwen2,
  title={Qwen2. 5-vl technical report},
  author={Bai, Shuai and Chen, Keqin and Liu, Xuejing and Wang, Jialin and Ge, Wenbin and Song, Sibo and Dang, Kai and Wang, Peng and Wang, Shijie and Tang, Jun and others},
  journal={arXiv preprint},
  year={2025}
}

@article{wang2025echoshot,
  title={EchoShot: Multi-Shot Portrait Video Generation},
  author={Wang, Jiahao and Sheng, Hualian and Cai, Sijia and Zhang, Weizhan and Yan, Caixia and Feng, Yachuang and Deng, Bing and Ye, Jieping},
  journal={NeurIPS},
  year={2025}
}

@article{song2025history,
  title={History-guided video diffusion},
  author={Song, Kiwhan and Chen, Boyuan and Simchowitz, Max and Du, Yilun and Tedrake, Russ and Sitzmann, Vincent},
  journal={ICML},
  year={2025}
}

@inproceedings{lu2025skald,
  title={SKALD: Learning-Based Shot Assembly for Coherent Multi-Shot Video Creation},
  author={Lu, Chen-Yi and Tanjim, Md Mehrab and Dasgupta, Ishita and Sarkhel, Somdeb and Wu, Gang and Mitra, Saayan and Chaterji, Somali},
  booktitle={ICCV},
  pages={17859--17868},
  year={2025}
}

@inproceedings{chuwan,
  title={Wan-Move: Motion-controllable Video Generation via Latent Trajectory Guidance},
  author={Chu, Ruihang and He, Yefei and Chen, Zhekai and Zhang, Shiwei and Xu, Xiaogang and Xia, Bin and Wang, Dingdong and Yi, Hongwei and Liu, Xihui and Zhao, Hengshuang and others},
  booktitle={NeurIPS},
  year={2025}
}

@article{chen2025talkcuts,
  title={TalkCuts: A Large-Scale Dataset for Multi-Shot Human Speech Video Generation},
  author={Chen, Jiaben and Wang, Zixin and Zeng, Ailing and Fu, Yang and Yu, Xueyang and Cen, Siyuan and Tanke, Julian and Chen, Yihang and Saito, Koichi and Mitsufuji, Yuki and others},
  journal={NeurIPS},
  year={2025}
}

@article{liao2025thinking,
  title={Thinking with Camera: A Unified Multimodal Model for Camera-Centric Understanding and Generation},
  author={Liao, Kang and Wu, Size and Wu, Zhonghua and Jin, Linyi and Wang, Chao and Wang, Yikai and Wang, Fei and Li, Wei and Loy, Chen Change},
  journal={arXiv preprint},
  year={2025}
}

@article{wang2025cinetechbench,
  title={CineTechBench: A Benchmark for Cinematographic Technique Understanding and Generation},
  author={Wang, Xinran and Xu, Songyu and Shan, Xiangxuan and Zhang, Yuxuan and Diao, Muxi and Duan, Xueyan and Huang, Yanhua and Liang, Kongming and Ma, Zhanyu},
  journal={arXiv preprint},
  year={2025}
}

@article{liu2025shotbench,
  title={ShotBench: Expert-Level Cinematic Understanding in Vision-Language Models},
  author={Liu, Hongbo and He, Jingwen and Jin, Yi and Zheng, Dian and Dong, Yuhao and Zhang, Fan and Huang, Ziqi and He, Yinan and Li, Yangguang and Chen, Weichao and others},
  journal={arXiv preprint},
  year={2025}
}

@article{lin2025towards,
  title={Towards Understanding Camera Motions in Any Video},
  author={Lin, Zhiqiu and Cen, Siyuan and Jiang, Daniel and Karhade, Jay and Wang, Hewei and Mitra, Chancharik and Ling, Tiffany and Huang, Yuhan and Liu, Sifan and Chen, Mingyu and others},
  journal={arXiv preprint},
  year={2025}
}

@article{bansal2024talc,
  title={TALC: Time-Aligned Captions for Multi-Scene Text-to-Video Generation},
  author={Bansal, Hritik and Bitton, Yonatan and Yarom, Michal and Szpektor, Idan and Grover, Aditya and Chang, Kai-Wei},
  journal={arXiv preprint},
  year={2024}
}

@article{bar2024lumiere,
  title={Lumiere: A space-time diffusion model for video generation},
  author={Bar-Tal, Omer and Chefer, Hila and Tov, Omer and Herrmann, Charles and Paiss, Roni and Zada, Shiran and Ephrat, Ariel and Hur, Junhwa and Liu, Guanghui and Raj, Amit and others},
  journal={arXiv preprint},
  year={2024}
}

@inproceedings{huang2024vbench,
  title={Vbench: Comprehensive benchmark suite for video generative models},
  author={Huang, Ziqi and He, Yinan and Yu, Jiashuo and Zhang, Fan and Si, Chenyang and Jiang, Yuming and Zhang, Yuanhan and Wu, Tianxing and Jin, Qingyang and Chanpaisit, Nattapol and others},
  booktitle={CVPR},
  pages={21807--21818},
  year={2024}
}

@article{kong2024hunyuanvideo,
  title={Hunyuanvideo: A systematic framework for large video generative models},
  author={Kong, Weijie and Tian, Qi and Zhang, Zijian and Min, Rox and Dai, Zuozhuo and Zhou, Jin and Xiong, Jiangfeng and Li, Xin and Wu, Bo and Zhang, Jianwei and others},
  journal={arXiv preprint},
  year={2024}
}

@misc{genmo2024mochi,
  title={Mochi 1},
  author={Genmo Team},
  year={2024},
  publisher = {GitHub},
  journal = {GitHub repository},
  howpublished={\url{https://github.com/genmoai/models}}
}

@inproceedings{esser2024scaling,
  title={Scaling rectified flow transformers for high-resolution image synthesis},
  author={Esser, Patrick and Kulal, Sumith and Blattmann, Andreas and Entezari, Rahim and M{\"u}ller, Jonas and Saini, Harry and Levi, Yam and Lorenz, Dominik and Sauer, Axel and Boesel, Frederic and others},
  booktitle={ICML},
  year={2024}
}

@article{zhang2025packing,
  title={Packing input frame context in next-frame prediction models for video generation},
  author={Zhang, Lvmin and Agrawala, Maneesh},
  journal={arXiv preprint},
  year={2025}
}

@inproceedings{long2024videostudio,
  title={VideoStudio: Generating Consistent-Content and Multi-Scene Videos},
  author={Long, Fuchen and Qiu, Zhaofan and Yao, Ting and Mei, Tao},
  booktitle={ECCV},
  pages={468--485},
  year={2024},
  organization={Springer}
}

@article{atzmon2024multi,
  title={Multi-Shot Character Consistency for Text-to-Video Generation},
  author={Atzmon, Yuval and Gal, Rinon and Tewel, Yoad and Kasten, Yoni and Chechik, Gal},
  journal={arXiv preprint},
  year={2024}
}

@article{zheng2024videogen,
  title={VideoGen-of-Thought: A Collaborative Framework for Multi-Shot Video Generation},
  author={Zheng, Mingzhe and Xu, Yongqi and Huang, Haojian and Ma, Xuran and Liu, Yexin and Shu, Wenjie and Pang, Yatian and Tang, Feilong and Chen, Qifeng and Yang, Harry and others},
  journal={arXiv preprint},
  year={2024}
}

@article{zhao2024moviedreamer,
  title={Moviedreamer: Hierarchical generation for coherent long visual sequence},
  author={Zhao, Canyu and Liu, Mingyu and Wang, Wen and Chen, Weihua and Wang, Fan and Chen, Hao and Zhang, Bo and Shen, Chunhua},
  journal={arXiv preprint},
  year={2024}
}

@article{luo2023videofusion,
  title={Videofusion: Decomposed diffusion models for high-quality video generation},
  author={Luo, Zhengxiong and Chen, Dayou and Zhang, Yingya and Huang, Yan and Wang, Liang and Shen, Yujun and Zhao, Deli and Zhou, Jingren and Tan, Tieniu},
  journal={arXiv preprint},
  year={2023}
}

@article{wang2024lavie,
  title={Lavie: High-quality video generation with cascaded latent diffusion models},
  author={Wang, Yaohui and Chen, Xinyuan and Ma, Xin and Zhou, Shangchen and Huang, Ziqi and Wang, Yi and Yang, Ceyuan and He, Yinan and Yu, Jiashuo and Yang, Peiqing and others},
  journal={International Journal of Computer Vision},
  pages={1--20},
  year={2024},
  publisher={Springer}
}

@inproceedings{chen2024livephoto,
  title={Livephoto: Real image animation with text-guided motion control},
  author={Chen, Xi and Liu, Zhiheng and Chen, Mengting and Feng, Yutong and Liu, Yu and Shen, Yujun and Zhao, Hengshuang},
  booktitle={ECCV},
  pages={475--491},
  year={2024},
  organization={Springer}
}

@article{zhang2024show,
  title={Show-1: Marrying pixel and latent diffusion models for text-to-video generation},
  author={Zhang, David Junhao and Wu, Jay Zhangjie and Liu, Jia-Wei and Zhao, Rui and Ran, Lingmin and Gu, Yuchao and Gao, Difei and Shou, Mike Zheng},
  journal={IJCV},
  pages={1--15},
  year={2024},
  publisher={Springer}
}

@inproceedings{kara2025shotadapter,
  title={ShotAdapter: Text-to-Multi-Shot Video Generation with Diffusion Models},
  author={Kara, Ozgur and Singh, Krishna Kumar and Liu, Feng and Ceylan, Duygu and Rehg, James M and Hinz, Tobias},
  booktitle={CVPR},
  pages={28405--28415},
  year={2025}
}

@article{guo2025long,
  title={Long context tuning for video generation},
  author={Guo, Yuwei and Yang, Ceyuan and Yang, Ziyan and Ma, Zhibei and Lin, Zhijie and Yang, Zhenheng and Lin, Dahua and Jiang, Lu},
  journal={ICCV},
  year={2025}
}

@inproceedings{wu2025mind,
  title={Mind the time: Temporally-controlled multi-event video generation},
  author={Wu, Ziyi and Siarohin, Aliaksandr and Menapace, Willi and Skorokhodov, Ivan and Fang, Yuwei and Chordia, Varnith and Gilitschenski, Igor and Tulyakov, Sergey},
  booktitle={CVPR},
  pages={23989--24000},
  year={2025}
}

@inproceedings{qi2025mask,
  title={Mask\^{} 2DiT: Dual Mask-based Diffusion Transformer for Multi-Scene Long Video Generation},
  author={Qi, Tianhao and Yuan, Jianlong and Feng, Wanquan and Fang, Shancheng and Liu, Jiawei and Zhou, SiYu and He, Qian and Xie, Hongtao and Zhang, Yongdong},
  booktitle={CVPR},
  pages={18837--18846},
  year={2025}
}

@inproceedings{shi2025motionstone,
  title={Motionstone: Decoupled motion intensity modulation with diffusion transformer for image-to-video generation},
  author={Shi, Shuwei and Gong, Biao and Chen, Xi and Zheng, Dandan and Tan, Shuai and Yang, Zizheng and Li, Yuyuan and He, Jingwen and Zheng, Kecheng and Chen, Jingdong and others},
  booktitle={CVPR},
  pages={22864--22874},
  year={2025}
}

@article{hacohen2024ltx,
  title={Ltx-video: Realtime video latent diffusion},
  author={HaCohen, Yoav and Chiprut, Nisan and Brazowski, Benny and Shalem, Daniel and Moshe, Dudu and Richardson, Eitan and Levin, Eran and Shiran, Guy and Zabari, Nir and Gordon, Ori and others},
  journal={arXiv preprint},
  year={2024}
}

@article{xie2024dreamfactory,
  title={Dreamfactory: Pioneering multi-scene long video generation with a multi-agent framework},
  author={Xie, Zhifei and Tang, Daniel and Tan, Dingwei and Klein, Jacques and Bissyand, Tegawend F and Ezzini, Saad},
  journal={arXiv preprint },
  year={2024}
}

@article{seawead2025seaweed,
  title={Seaweed-7b: Cost-effective training of video generation foundation model},
  author={Seawead, Team and Yang, Ceyuan and Lin, Zhijie and Zhao, Yang and Lin, Shanchuan and Ma, Zhibei and Guo, Haoyuan and Chen, Hao and Qi, Lu and Wang, Sen and others},
  journal={arXiv preprint },
  year={2025}
}

@inproceedings{guo2025keyframe,
  title={Keyframe-Guided Creative Video Inpainting},
  author={Guo, Yuwei and Yang, Ceyuan and Rao, Anyi and Meng, Chenlin and Bar-Tal, Omer and Ding, Shuangrui and Agrawala, Maneesh and Lin, Dahua and Dai, Bo},
  booktitle={CVPR},
  pages={13009--13020},
  year={2025}
}

@article{wang2025frame,
  title={Frame In-N-Out: Unbounded Controllable Image-to-Video Generation},
  author={Wang, Boyang and Chen, Xuweiyi and Gadelha, Matheus and Cheng, Zezhou},
  journal={NeurIPS},
  year={2025}
}

@inproceedings{li2025realcam,
  title={Realcam-i2v: Real-world image-to-video generation with interactive complex camera control},
  author={Li, Teng and Zheng, Guangcong and Jiang, Rui and Zhan, Shuigen and Wu, Tao and Lu, Yehao and Lin, Yining and Deng, Chuanyun and Xiong, Yepan and Chen, Min and others},
  booktitle={ICCV},
  pages={28785--28796},
  year={2025}
}

@inproceedings{lin2025stiv,
  title={Stiv: Scalable text and image conditioned video generation},
  author={Lin, Zongyu and Liu, Wei and Chen, Chen and Lu, Jiasen and Hu, Wenze and Fu, Tsu-Jui and Allardice, Jesse and Lai, Zhengfeng and Song, Liangchen and Zhang, Bowen and others},
  booktitle={ICCV},
  pages={16249--16259},
  year={2025}
}

@article{wang2024dreamrunner,
  title={DreamRunner: Fine-Grained Storytelling Video Generation with Retrieval-Augmented Motion Adaptation},
  author={Wang, Zun and Li, Jialu and Lin, Han and Yoon, Jaehong and Bansal, Mohit},
  journal={arXiv preprint },
  year={2024}
}

@article{hu2024storyagent,
  title={StoryAgent: Customized Storytelling Video Generation via Multi-Agent Collaboration},
  author={Hu, Panwen and Jiang, Jin and Chen, Jianqi and Han, Mingfei and Liao, Shengcai and Chang, Xiaojun and Liang, Xiaodan},
  journal={arXiv preprint },
  year={2024}
}

@article{huang2025conceptmaster,
  title={ConceptMaster: Multi-Concept Video Customization on Diffusion Transformer Models Without Test-Time Tuning},
  author={Huang, Yuzhou and Yuan, Ziyang and Liu, Quande and Wang, Qiulin and Wang, Xintao and Zhang, Ruimao and Wan, Pengfei and Zhang, Di and Gai, Kun},
  journal={arXiv preprint },
  year={2025}
}

@inproceedings{jiang2024videobooth,
  title={Videobooth: Diffusion-based video generation with image prompts},
  author={Jiang, Yuming and Wu, Tianxing and Yang, Shuai and Si, Chenyang and Lin, Dahua and Qiao, Yu and Loy, Chen Change and Liu, Ziwei},
  booktitle={CVPR},
  pages={6689--6700},
  year={2024}
}

@inproceedings{xing2024dynamicrafter,
  title={Dynamicrafter: Animating open-domain images with video diffusion priors},
  author={Xing, Jinbo and Xia, Menghan and Zhang, Yong and Chen, Haoxin and Yu, Wangbo and Liu, Hanyuan and Liu, Gongye and Wang, Xintao and Shan, Ying and Wong, Tien-Tsin},
  booktitle={ECCV},
  pages={399--417},
  year={2024},
  organization={Springer}
}

@article{lipman2022flow,
  title={Flow matching for generative modeling},
  author={Lipman, Yaron and Chen, Ricky TQ and Ben-Hamu, Heli and Nickel, Maximilian and Le, Matt},
  journal={arXiv preprint },
  year={2022}
}

@article{liu2022flow,
  title={Flow straight and fast: Learning to generate and transfer data with rectified flow},
  author={Liu, Xingchao and Gong, Chengyue and Liu, Qiang},
  journal={arXiv preprint },
  year={2022}
}

@inproceedings{rombach2022high,
  title={High-resolution image synthesis with latent diffusion models},
  author={Rombach, Robin and Blattmann, Andreas and Lorenz, Dominik and Esser, Patrick and Ommer, Bj{\"o}rn},
  booktitle={CVPR},
  pages={10684--10695},
  year={2022}
}

@article{wan2025wan,
  title={Wan: Open and advanced large-scale video generative models},
  author={Wan, Team and Wang, Ang and Ai, Baole and Wen, Bin and Mao, Chaojie and Xie, Chen-Wei and Chen, Di and Yu, Feiwu and Zhao, Haiming and Yang, Jianxiao and others},
  journal={arXiv preprint },
  year={2025}
}

@article{chasanis2008scene,
  title={Scene detection in videos using shot clustering and sequence alignment},
  author={Chasanis, Vasileios T and Likas, Aristidis C and Galatsanos, Nikolaos P},
  journal={IEEE transactions on multimedia},
  volume={11},
  number={1},
  pages={89--100},
  year={2008},
  publisher={IEEE}
}

\clearpage
\newpage
\beginappendix

\section{Additional Training Details}
\label{sec:app_training}

This section provides expanded details on the training formulation used in our model, including the unified three-shot training setup and the construction of frame-level pseudo-labels.  
These details complement~\Cref{sec:frame_selection} and~\Cref{sec:training} of the main paper.

\subsection{Unified Three-Shot Training}
\label{sec:app_unified}
As discussed in~\Cref{sec:data_process} of the main paper, the dataset contains videos with varying numbers of shots, with two-shot sequences being the most common and three-shot sequences relatively fewer.  
Training directly on sequences of non-uniform length leads to unstable optimization.  
To mitigate this, we unify all training samples into a \emph{three-shot format} by synthesizing an additional shot for two-shot videos.

\noindent\textbf{Synthetic shot construction.}
Given a two-shot sequence $(S_{\mathrm{first}}, S_{\mathrm{last}})$, we create a synthetic shot $S_{\mathrm{syn}}$ using one of:
\begin{enumerate}[label=(\roman*), leftmargin=18pt]
    \item Cross-video insertion: inserting a shot randomly sampled from another video.
    \item Augmented-first-shot variant: applying spatial or color transformations to $S_{\mathrm{first}}$.
\end{enumerate}

This results in synthetic triplets that, for each sample, take one of the two forms:
\begin{equation}
\big(S_{\mathrm{first}}, S_{\mathrm{syn}}, S_{\mathrm{last}}\big) \quad \text{or} \quad
\big(S_{\mathrm{syn}}, S_{\mathrm{first}}, S_{\mathrm{last}}\big),
\end{equation}
while the real triplets are represented in the structure $(S_{\mathrm{first}}, S_{\mathrm{second}}, S_{\mathrm{last}})$.  
In all cases, $S_{\mathrm{last}}$ serves as the prediction target.

\noindent\textbf{Training objective.}
The model is trained to generate the final shot $S_{\mathrm{last}}$ conditioned on the first two shots and its caption $C_{\mathrm{last}}$:
\begin{equation}
\label{eq:shot_loss_app}
\mathcal{L}_{\mathrm{shot}}
= \mathbb{E}\!\left[
\mathcal{L}_{\mathrm{diff}}\big(
\mathcal{G}(S_{\mathrm{first}},\, S_{\mathrm{syn}/second},\, C_{\mathrm{last}}),\;
S_{\mathrm{last}}
\big)
\right],
\end{equation}
where $\mathcal{L}_{\mathrm{diff}}$ denotes a rectified-flow diffusion loss~\citep{lipman2022flow,liu2022flow,esser2024scaling}.  
This unified formulation standardizes all training samples to a consistent three-shot structure and enables unified three-shot training, improving optimization stability.

\subsection{Frame Relevance Pseudo-Labels}
\label{sec:app_pseudolabels}

To assist the learning of the frame relevance scores $\mathbf{S}$, we construct frame-level pseudo-labels $\mathbf{y}=\{y_r\}_{r=1}^{F}$ that approximate the relevance of each historical frame in $\mathbf{M}$ to the target shot.
The pseudo-labels incorporate both real and synthetic frames introduced in~\Cref{sec:app_unified}.

\noindent\textbf{Real historical frames.}
For frames originating from real historical shots, we compute cosine similarity between each historical frame and the target shot using DINOv2~\citep{oquab2023dinov2} and CLIP~\citep{radford2021learning} embeddings, producing a scalar relevance score.  
These pseudo-labels help the Frame Selection module prioritize visually and semantically aligned frames while down-weighting irrelevant ones.

\noindent\textbf{Synthetic frames.}
Frames from synthetic shots introduced in~\Cref{sec:app_unified} are assigned coarse relevance labels:  
$y_r = -1$ for randomly inserted shots to indicate clear irrelevance, and  
$y_r = 0$ for augmented-first-shot variants to reflect partial relevance.  
These labels explicitly guide the selector to down-weight non-informative or misleading frames.

\noindent\textbf{Supervision loss.}
The predicted relevance scores $\mathbf{S}$ are supervised using a regression loss:
\begin{equation}
\label{eq:sel_loss}
\mathcal{L}_{\mathrm{sel}} = \frac{1}{F}\sum_{r=1}^{F}(s_r - y_r)^2,
\end{equation}
where $s_r = \mathbf{S}[r]$ denotes the predicted relevance score for the $r^{th}$ historical frame.  
The full training objective is given by:
\begin{equation}
\label{eq:train_loss}
\mathcal{L}_{\mathrm{train}} = \mathcal{L}_{\mathrm{shot}} + \lambda\,\mathcal{L}_{\mathrm{sel}},
\end{equation}
with $\lambda$ controlling the weight of the selector supervision loss.  
This joint optimization encourages the model to identify informative context frames while maintaining high-fidelity generation.

\section{Additional Details on Evaluation Benchmark}
\label{sec:eval_data}

We construct a human-centric benchmark for both T2MSV and I2MSV to evaluate multi-shot video generation under realistic narrative conditions.  
As introduced in~\Cref{sec:data_process} of the main paper, each shot is paired with a referential caption following a progressive narrative flow, reflecting real-world storytelling.
To comprehensively evaluate MSV performance, the benchmark spans three canonical multi-shot storytelling patterns:

\begin{enumerate}
    \item \textbf{Main-subject consistency.}  
    Multiple shots focus on the same character(s), who may appear in different environments or perform different actions.  
    This pattern evaluates the model’s ability to preserve identity under various cross-shot changes.

    \item \textbf{Insert-and-recall with an intervening shot.}  
    A shot introducing a new scene, such as an environment-only view or a new character, is inserted mid-sequence, after which the narrative returns to the primary subject(s) and later revisits the intervening shot.
    This pattern stresses the model’s ability to maintain long-range memory and remain robust to temporal distractors.

    \item \textbf{Composable generation.}  
    Characters introduced separately in earlier shots are composed together in later shots.  
    This tests whether the model can correctly integrate multiple narrative threads into a coherent shared scene.
\end{enumerate}

In total, we curate 64 six-shot test cases for T2MSV and 64 six-shot test cases for I2MSV, covering a diverse range of subjects, environments, and complex cross-shot relationships, thereby ensuring comprehensive MSV performance evaluation.
More examples are provided in our \projectpage.

\section{Additional Qualitative Results}

Generating \textit{coherent multi-shot videos} that faithfully follow narrative captions is essential for real-world storytelling.  
Here, we analyze our model from three perspectives, using examples from the main paper to illustrate its ability to maintain continuity across shots.
Additional video qualitative results are available our \projectpage.

\noindent\textbf{Identity consistency.}  
Our model preserves character identity across long-range shots under diverse variations.  
In the $1^{st}$ example of~\Cref{fig:teaser} in the main paper, the same subject remains consistent across changes in viewpoint (Shots~4, 5) and actions (Shots~1, 3, 8).  
This illustrates the effectiveness of our adaptive memory in maintaining stable long-range identity cues.

\noindent\textbf{Background details.}  
Beyond character fidelity, our model maintains consistent background details across shots, enabling spatially coherent story progression.  
In the $2^{nd}$ example of~\Cref{fig:teaser} in the main paper, fine-grained elements such as plants and fences remain aligned from Shot~1 to Shot~7 despite large cross-shot dynamics.  
Similarly, in the $3^{rd}$ example, the red flowers reappear consistently across Shots~1, 4, 5, 6, 7, and 9, demonstrating the model’s ability to preserve scene layout and spatial structure.

\noindent\textbf{Reappearance and composition.}  
Realistic narratives often involve disappear–reappear patterns and the merging of multiple narrative threads through composable generation.  
Our model effectively recalls characters or environments that reemerge after several intervening shots, \eg, Shots~4 and 9, or Shots~2 and 6 in the $2^{nd}$ example of~\Cref{fig:teaser} in the main paper.  
Furthermore, in Shot~7 of the same example, the woman from Shot~1 and the man from Shot~4 appear together, demonstrating the model’s capacity to unify distinct visual narratives into a coherent multi-subject scene.

\end{document}